\documentclass[final]{cvpr}

\usepackage{times}
\usepackage{epsfig}
\usepackage{graphicx}
\usepackage{amsmath}
\usepackage{amssymb}

\usepackage{color}
\usepackage{colortbl}

\makeatletter
\@namedef{ver@everyshi.sty}{}
\makeatother
\usepackage{tikz,pgfplots}

\definecolor{note}{rgb}{0.1,0.1,1}
\definecolor{remove}{rgb}{0.7,0.1,0.1}
\definecolor{TODO}{rgb}{0.9,0.5,0.1}

\definecolor{rowblue}{rgb}{0.9,0.95,1}
\definecolor{rowgray}{rgb}{0.95,0.95,0.9375}

\definecolor{blue}{rgb}{0.184313725,	0.333333333,	0.592156863}
\definecolor{red}{rgb}{0.517647059,	0.235294118,	0.047058824}
\definecolor{green}{rgb}{0.219607843,	0.341176471,	0.137254902}
\definecolor{yellow}{rgb}{0.498039216,	0.376470588,	0}
\definecolor{gray}{rgb}{0.40784,	0.40392,	0.40392}
\definecolor{magenta}{rgb}{0.184313725,	0.333333333,	0.592156863}

\usepackage[pagebackref=true,breaklinks=true,colorlinks,bookmarks=false]{hyperref}

\def\cvprPaperID{1159} 
\def\confYear{CVPR 2021}

\begin{document}

\title{Depth from Camera Motion and Object Detection}

\author{Brent A. Griffin \\
	University of Michigan\\
	{\tt\small griffb@umich.edu}
	\and
	Jason J. Corso \\ Stevens Institute of Technology \\ {\tt\small jcorso@stevens.edu}
}

\maketitle

\begin{abstract}
	
This paper addresses the problem of learning to estimate the depth of detected objects given some measurement of camera motion (e.g., from robot kinematics or vehicle odometry). We achieve this by 1) designing a recurrent neural network (DBox) that estimates the depth of objects using a generalized representation of bounding boxes and uncalibrated camera movement and 2) introducing the Object Depth via Motion and Detection Dataset (ODMD). ODMD training data are extensible and configurable, and the ODMD benchmark includes 21,600 examples across four validation and test sets. These sets include mobile robot experiments using an end-effector camera to locate objects from the YCB dataset and examples with perturbations added to camera motion or bounding box data. In addition to the ODMD benchmark, we evaluate DBox in other monocular application domains, achieving state-of-the-art results on existing driving and robotics benchmarks and estimating the depth of objects using a camera phone.

	
\end{abstract}

\section{Introduction}

\begin{figure} [t]
	\centering
	\includegraphics[width=0.435 \textwidth]{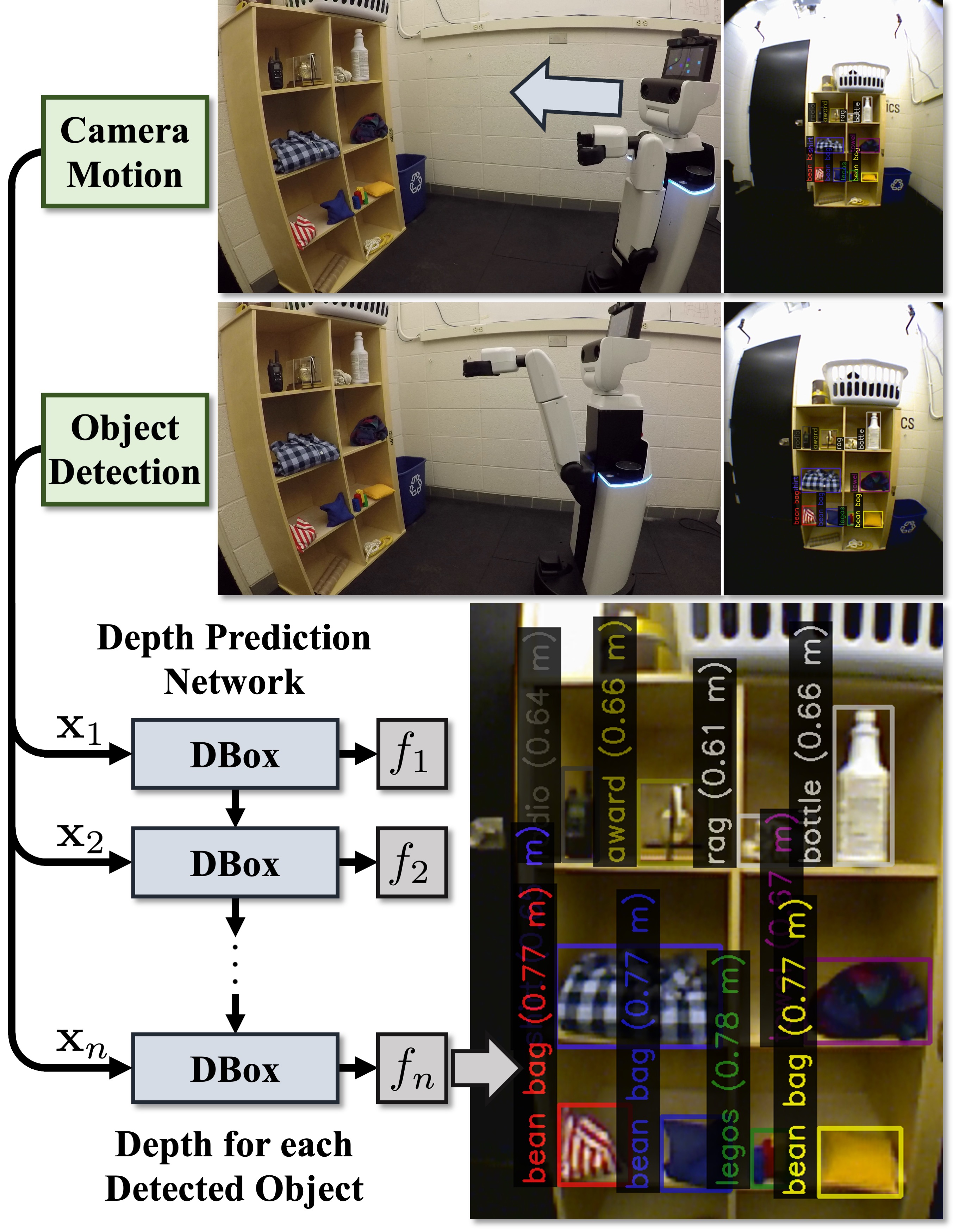}
	\caption{ \textbf{Depth from Camera Motion and Object Detection}.
		Object detectors can reliably place bounding boxes on target objects in a variety of settings.
		Given a sequence of bounding boxes and camera movement distances between observations (e.g., from robot kinematics, top), our network (DBox) estimates each object's depth (bottom).
		This result is from the ODMD Robot Set.
	}
	\label{fig:overview}
\end{figure}

With the progression of high-quality datasets and subsequent methods, our community has seen remarkable advances in image segmentation \cite{KiEtAl19,BoEtAl17}, video object segmentation \cite{OhEtAl19,DAVIS}, and object detection \cite{CaEtAl20,LVIS,MSCOCO}, sometimes with a focus on a specific application like driving \cite{City,KITTI,SYNTHIA}.
However, applications like autonomous vehicles and robotics require a three-dimensional (3D) understanding of the environment, so they frequently rely on 3D sensors (e.g., LiDAR \cite{GaEtAl20} or RGBD cameras \cite{FeLa19}).
Although 3D sensors are great for identifying free space and motion planning, classifying and understanding raw 3D data is a challenging and ongoing area of research \cite{KhEtAl19,SuSh19,LiuEtAl19,pointnet}.
On the other hand, RGB cameras are inexpensive, ubiquitous, and interpretable by countless vision methods.



To bridge the gap between 3D applications and progress in video object segmentation, in recent work \cite{GrFlCo20}, we developed a method of video object segmentation-based visual servo control, object depth estimation, and mobile robot grasping using a single RGB camera.
For object depth estimation, specifically, we used the optical expansion \cite{It51,SwGo86} of segmentation masks with $z$-axis camera motion to analytically solve for depth \cite[VOS-DE (20)]{GrFlCo20}.
In subsequent work \cite{GrCo20}, we introduced the first learning-based method (ODN) and benchmark dataset for estimating Object Depth via Motion and Segmentation (ODMS), which includes test sets in robotics and driving.
From the ODMS benchmark, ODN improves accuracy over VOS-DE in multiple domains, especially those with segmentation errors.



Motivated by these developments \cite{GrFlCo20,GrCo20}, this paper addresses the problem of estimating the depth of objects using uncalibrated camera motion and bounding boxes from object detection (see Figure~\ref{fig:overview}), which has many advantages.
First, a bounding box has only four parameters, can be processed quickly with few resources, and has less domain-specific features than an RGB image or segmentation mask.
Second, movement is already measured on most autonomous hardware platforms and, even if not measured, structure from motion is plausible to recover camera motion \cite{KaGaBa19,MuTa17,ScFr16}.
Third, as we show with a pinhole camera and box-based model (see Figure~\ref{fig:3D_box}) and in experiments, we can use $x$-, $y$-, or $z$-axis camera motion to estimate depth using optical expansion, motion parallax \cite{Fe72,RoGr79}, or both.
Finally, our detection-based methods can support more applications by using boxes \textit{or} segmentation masks, which we demonstrate in multiple domains with state-of-the-art results on the segmentation-based ODMS benchmark \cite{GrCo20}.


The first contribution of our paper is deriving an analytical model and corresponding solution (Box$_{\text{LS}}$) for uncalibrated motion and detection-based depth estimation in Section~\ref{sec:model}.
To the best of our knowledge, this is the first model or solution in this new problem space.
Furthermore, Box$_{\text{LS}}$ achieves the best analytical result on the ODMS benchmark.

A second contribution is developing a recurrent neural network (RNN) to predict {\bf D}epth from motion and bounding {\bf Box}es (DBox) in Section~\ref{sec:dbox}.
DBox sequentially processes observations and uses our normalized and dimensionless input-loss formulation, which improves performance across domains with different movement distances and camera parameters.
Thus, using a single DBox network, we achieve the best Robot, Driving, and overall result on the ODMS benchmark and estimate depth from a camera phone.\footnote{Supplementary video: \url{https://youtu.be/GruhbdJ2l7k}}


Inspired by ODMS \cite{GrCo20}, a final contribution of our paper is the \textbf{O}bject \textbf{D}epth via \textbf{M}otion and \textbf{D}etection (ODMD) dataset in Section~\ref{sec:odmd}.\footnote{Dataset website: \url{https://github.com/griffbr/ODMD}}
ODMD is the first dataset for motion and detection-based depth estimation, which enables learning-based methods in this new problem space.
ODMD data consist of a series of bounding boxes, $x,y,z$ camera movement distances, and ground truth object depth.

For ODMD training, we continuously generate synthetic examples with random movements, depths, object sizes, and three types of perturbations typical of camera motion and object detection errors.
As we will show, training with perturbations improves end performance in real applications.
Furthermore, ODMD's distance- and box-based inputs are 1) simple, so we can generate over 300,000 training examples per second, and 2) general, so we can transfer from synthetic training data to many application domains.


Finally, for an ODMD evaluation benchmark, we create four validation and test sets with 21,600 examples, including mobile robot experiments locating YCB objects \cite{YCB}.

\section{Related Work}

\noindent \textbf{Object Detection} predicts a set of bounding boxes and category labels for objects of interest. 
Many object detectors operate using regression and classification over a set of region proposals \cite{CaVa18,ReEtAl15}, anchors \cite{LiEtAl17b}, or window centers \cite{TiEtAl19}.
Other detectors treat detection as a single regression problem \cite{ReEtAl16} or, more recently, use a transformer architecture \cite{VaEtAl17} to directly predict all detections in parallel \cite{CaEtAl20}.
Given the utility of locating objects in RGB images, detection supports many downstream vision tasks such as segmentation \cite{maskrcnn}, 3D shape prediction \cite{GkMaJo19}, object pose estimation \cite{PaPaVi19}, and even single-view metrology \cite{ZhEtAl20}, to name but a few.

In this work, we provide depth for ``free" as an extension of object detection in mobile applications. 
We detect objects on a per-frame basis, then use sequences of bounding boxes with uncalibrated camera motion to find the depth of each object.
One benefit of our approach is that depth accuracy will improve with future detection methods.
For experiments, we use Faster R-CNN \cite{ReEtAl15}, which has had many improvements since its original publication, runs in real time, and is particularly accurate for small objects \cite{CaEtAl20}.
Specifically, we use the same Faster R-CNN configuration as Detectron2 \cite{detectron2} with ResNet~50 \cite{HeEtAl16} pre-trained on ImageNet \cite{ImageNet} and a FPN \cite{LiEtAl17} backbone trained on COCO \cite{MSCOCO},
which we then fine-tune for our validation and test set objects.

\vspace{2mm}
\noindent \textbf{Object Pose Estimation}, to its fullest extent, predicts the 3D position and 3D rotation of an object in camera-centered coordinates, which is useful for autonomous driving, augmented reality, and robot grasping \cite{SaEtAl20}.
Pose estimation methods for household objects typically use a single RGB image \cite{WaEtAl17,PaPaVi19}, RGBD image \cite{WaEtAl19}, or separate setting for each \cite{BrEtAl16,XiEtAl18}.
Alternatively, recent work \cite{LaEtAl20} uses multiple views to jointly predict multiple object poses, which achieves the best result on the YCB-Video dataset \cite{XiEtAl18}, T-LESS dataset \cite{HoEtAl17}, and BOP Challenge 2020 \cite{HoEtAl18}.

To predict object depth without scale ambiguity, many RGB-based pose estimation methods learn a prior on specific 3D object models \cite{YCB}.
However, innovations in robotics are easing this requirement.
Recent work \cite{PaPaVi20} uses a robot to collect data in cluttered scenes to add texture to simplified pose training models.
Other work \cite{DeEtAl20} uses a robot to interact with objects and generate new training data to improve its pose estimation. 
In our previous work \cite{GrFlCo20,GrCo20}, we use robot motion with segmentation to predict depth and grasp objects, which removes 3D models entirely.

In this work, we build off of these developments to improve RGB-based object depth estimation without any 3D model requirements.
Instead, we use depth cues based on uncalibrated camera movement and general object detection or segmentation.
A primary benefit of our approach is its generalization across domains, which we demonstrate in robotics, driving, and camera phone experiments.

\section{Depth from Camera Motion and \\Object Detection}

We design an RNN to predict {\bf D}epth from camera motion and bounding {\bf Box}es (DBox).
DBox sequentially processes each observation and uses optical expansion and motion parallax cues to make its final object depth prediction.
To train and evaluate DBox, we introduce the {\bf O}bject {\bf D}epth via {\bf M}otion and {\bf D}etection Dataset (ODMD).
First, in Section~\ref{sec:model}, we derive our motion and detection model with analytical solutions, which is the theoretical foundation for this work and informs our design decisions in the remaining paper.
Next, in Section~\ref{sec:dbox}, we detail DBox's input-loss formulation and architecture.
Finally, in Section~\ref{sec:odmd}, we explain ODMD's extensible training data, validation and test sets for evaluation, and DBox training configurations. 

\subsection{Depth from Motion and Detection Model}
\label{sec:model}



\noindent \textbf{Motion and Detection Inputs}. 
To find a detected object's depth, assume we are given a set of $n \geq 2$ observations 
$\mathbf{X} := \{\mathbf{x}_1, \mathbf{x}_2, \cdots, \mathbf{x}_n\}$,
where each observation $\mathbf{x}_i$ consists of a bounding box for the detected object and a corresponding camera position. Specifically,
\begin{align}
\mathbf{x}_i := \begin{bmatrix}
x_i, y_i, w_i, h_i,\mathbf{p}_i^\intercal
\end{bmatrix}^\intercal,
\label{eq:xi}
\end{align}
where $x_i, y_i, w_i, h_i$ denote the two image coordinates of the bounding box center, width, and height and 
\begin{align}
	\mathbf{p}_i := \begin{bmatrix} \text{C}_{Xi}, \text{C}_{Yi}, \text{C}_{Zi} \end{bmatrix}^\intercal
	\label{eq:pi}
\end{align}
is the relative camera position for each observation $\mathbf{x}_i \in \mathbb{R}^7$.
Notably, we align the axes of $\mathbf{p}_i$ with the camera coordinate frame and the model is most accurate without camera rotation, but the absolute position of $\mathbf{p}_i$ is inconsequential.
Observations can be collected as a set of images or by video.


\vspace{2mm}
\noindent \textbf{Camera Model}. 
To infer 3D information from 2D detection, we relate an object's bounding box image points to 3D camera-frame coordinates using the pinhole camera model
\begin{align}
\begin{bmatrix}
x \\ y
\end{bmatrix}
=
\begin{bmatrix}
f_x  & 0 & c_x \\ 0 & f_y & c_y
\end{bmatrix}
\begin{bmatrix}
\frac{X}{Z_i}, \frac{Y}{Z_i}, 1 \end{bmatrix}^\intercal,
\label{eq:proj}
\end{align}
where $f_x, f_y$ and $c_x, c_y$ are the camera's focal lengths and principal points and $X, Y$ correspond to image coordinates $x, y$ in the 3D camera frame at depth $Z_i$.
Notably, $Z_i$ is the distance along the optical axis (or depth) between the camera and the visible perimeter of the detected object.
To specify individual $x, y$ image coordinates we simplify \eqref{eq:proj} to
\begin{align}
x = \frac{f_x X}{Z_i} + c_x, ~ y = \frac{f_y Y}{Z_i} + c_y.
\label{eq:x}
\end{align}
Notably, although we include pinhole camera parameters in the model, we do not use them to solve depth at inference.

\begin{figure} [t!]
	\centering
	\includegraphics[width=0.415 \textwidth]{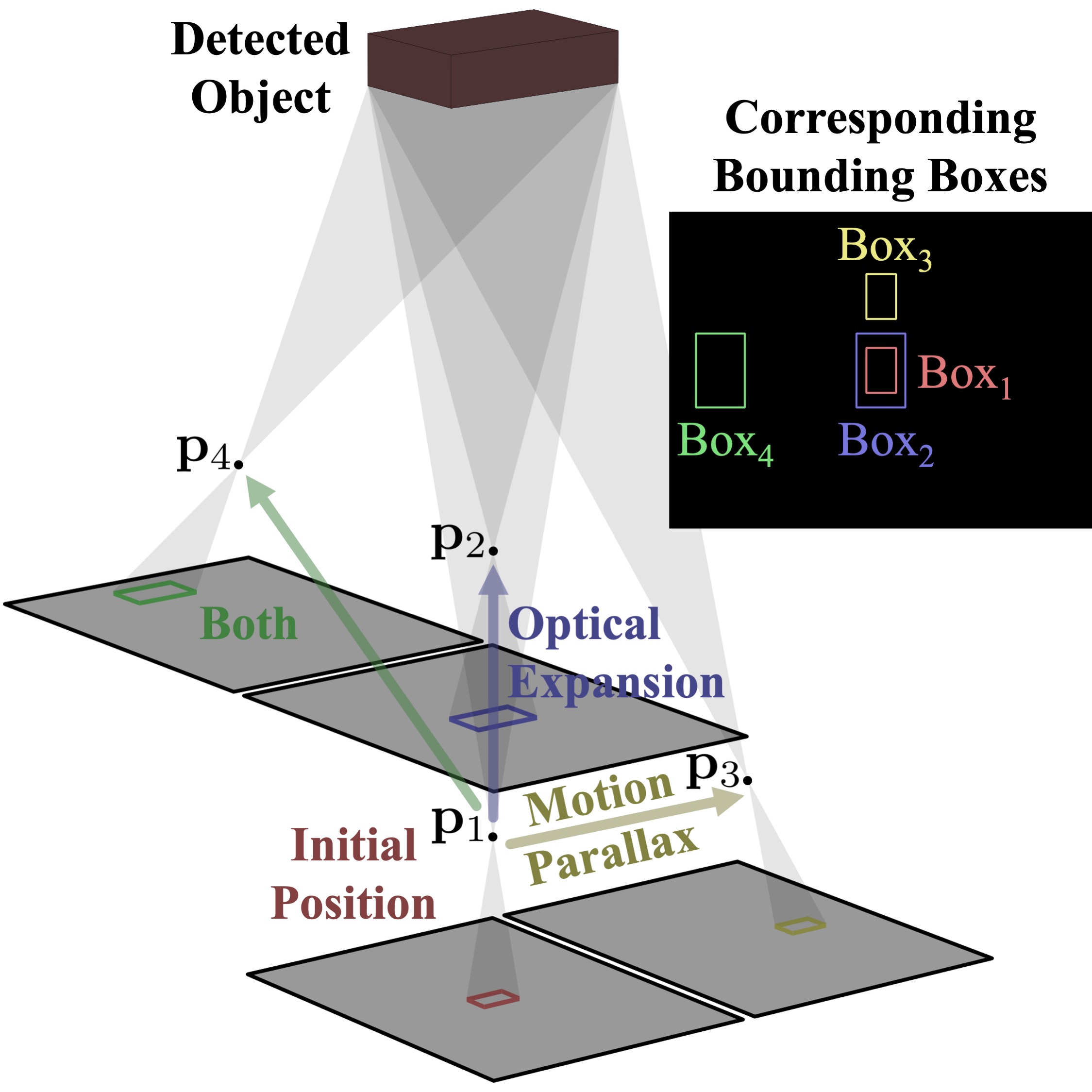}
	\caption{ \textbf{Motion and Detection Model}. 
		We show an object's bounding box (Box$_i$) for four pinhole camera positions ($\mathbf{p}_i$).
		Boxes scale inversely with object depth (optical expansion) and offset with lateral position (motion parallax).
		Using relative camera movements with these cues, we can solve for an object's depth.
	}
	\label{fig:3D_box}
\end{figure}

\vspace{2mm}
\noindent \textbf{Depth from Optical Expansion \& Detection}.
In an ideal model, we can find object depth using $z$-axis motion between observations and corresponding changes in bounding box scale (e.g., Box$_1$ to Box$_2$ in Figure~\ref{fig:3D_box}).
First, we use \eqref{eq:x} to relate bounding box width $w_i$ to 3D object width $W$ as 
\begin{align}
w_i = x_{\text{R}i} - x_{\text{L}i} =  \frac{f_x X_{\text{R}i}}{Z_i} + c_x  - \frac{f_x X_{\text{L}i}}{Z_i} - c_x = \frac{f_x W}{Z_i},
\label{eq:w}
\end{align}
where $x_{\text{R}i}, x_{\text{L}i}$ are the right and left box image coordinates with 3D coordinates $X_{\text{R}i}, X_{\text{L}i}$ respectively.
Object width $W$ is constant, so we use \eqref{eq:w} to relate two observations as 
\begin{align}
w_i Z_i = w_j Z_j = f_x W. \label{eq:fxw}
\end{align}
Notably, \eqref{eq:w}-\eqref{eq:fxw} also apply to object height $H$. Specifically,
$h_i = \frac{f_y H}{Z_i}$ from \eqref{eq:w} and $h_i Z_i = h_j Z_j = f_y H$ from \eqref{eq:fxw}.

To use the camera motion ($\mathbf{p}_i$), we note that changes in depth $Z_i$ between observations of a static object are only caused by changes in camera position $\text{C}_{Zi}$ \eqref{eq:pi}. Thus,
\begin{align}
Z_i + \text{C}_{Zi} = Z_j + \text{C}_{Zj} \implies Z_j = Z_i + \text{C}_{Zi} - \text{C}_{Zj}.
\label{eq:zj}
\end{align}

Finally, we use \eqref{eq:zj} in \eqref{eq:fxw} and solve for object depth $Z_i$ as
\begin{align}
w_i Z_i &~= w_j (Z_i + \text{C}_{Zi} - \text{C}_{Zj}) = f_x W  \nonumber \\
\implies Z_i &~= \frac{\text{C}_{Zj} - \text{C}_{Zi}}{1 - \big( \frac{w_i}{w_j} \big) }.
\label{eq:zi}
\end{align}
In \eqref{eq:zi}, we find $Z_i$ from two observations $i,j$ using optical expansion, i.e., the change in bounding box scale ($\frac{w_i}{w_j}$) relative to $z$-axis camera motion ($\text{C}_{Zj} - \text{C}_{Zi}$).
To measure the change in box scale using height, we replace $\frac{w_i}{w_j}$ with $\frac{h_i}{h_j}$.



\vspace{2mm}
\noindent \textbf{Depth from Motion Parallax \& Detection}. 
If there is $x$- or $y$-axis camera motion, we can solve for object depth using corresponding changes in bounding box location (e.g., Box$_1$ to Box$_3$ in Figure~\ref{fig:3D_box}).
For brevity, we provide this derivation and comparative results in the supplementary material.

\vspace{2mm}
\noindent \textbf{Using all Observations to Improve Depth}. 
In real applications, camera motion measurements and object detection will have errors.
Thus, we make detection-based depth estimation more robust by incorporating all $n$ observations.

For the $n$-observation solution, we reformulate \eqref{eq:zi} for each observation as 
$w_j Z_i - f_x W = w_j (\text{C}_{Zj} - \text{C}_{Zi})$ for width and $h_j Z_i - f_y H = h_j (\text{C}_{Zj} - \text{C}_{Zi})$ for height.
Now we can estimate $Z_i$ over $n$ observations in $\mathbf{A}\mathbf{x}=\mathbf{b}$ form as
\begin{align}
\begin{bmatrix}
w_1 & 1 & 0 \\ h_1 & 0 & 1 \\ w_2 & 1 & 0 \\ \vdots & \vdots & \vdots \\ h_n & 0 & 1
\end{bmatrix}
\begin{bmatrix}
\hat{Z}_i \\ -f_x \hat{W} \\ -f_y \hat{H}
\end{bmatrix}
=
\begin{bmatrix}
w_1 (\text{C}_{Z1} - \text{C}_{Zi}) \\ h_1 (\text{C}_{Z1} - \text{C}_{Zi}) \\ w_2 (\text{C}_{Z2} - \text{C}_{Zi}) \\ \vdots \\ h_n (\text{C}_{Zn} - \text{C}_{Zi})
\end{bmatrix}.
\label{eq:axb}
\end{align}
In this work, we solve for $\hat{Z}_i$ as a least-squares approximation of $Z_i$.
In Section~\ref{sec:results}, we refer to this solution as Box$_{\text{LS}}$.

\subsection{Depth from Motion and Detection Network}
\label{sec:dbox}

Considering the motion and detection model and Box$_{\text{LS}}$ solution in Section~\ref{sec:model}, we design DBox to use all $n$ observations for robustness and full $x,y,z$ camera motion to utilize both optical expansion and motion parallax cues.
Additionally, to improve performance across domains, we derive a normalized and dimensionless input-loss formulation. 

\vspace{2mm}
\noindent \textbf{Normalized Network Input}. 
As in Section~\ref{sec:model}, assume we have a set of $n \geq 2$ observations $\mathbf{X}$ to predict depth. 
We normalize the bounding box coordinates of each $\mathbf{x}_i$ \eqref{eq:xi} as 
\begin{align}
\mathbf{\bar{b}}_i := \begin{bmatrix}
	\frac{x_i}{W_I}, \frac{y_i}{H_I}, \frac{w_i}{W_I}, \frac{h_i}{H_I}
	\end{bmatrix}^\intercal,
\label{eq:bbar}
\end{align}
where $W_I$ and $H_I$ are the box image's width and height.
We normalize the camera position $\mathbf{p}_i$ \eqref{eq:pi} of each $\mathbf{x}_i$ as
\begin{align}
\mathbf{\bar{p}}_i :=
\frac{\mathbf{p}_i - \mathbf{p}_{i-1}}{\left\| \mathbf{p}_n - \mathbf{p}_1 \right\|},
\label{eq:pbar}
\end{align}
where $\left\| \mathbf{p}_n - \mathbf{p}_1 \right\|$ is the overall Euclidean camera movement range, $\mathbf{p}_i - \mathbf{p}_{i-1}$ is the incremental camera movement, and we set initial condition $\mathbf{p}_0 = \mathbf{p}_1 \implies \mathbf{\bar{p}}_1 = \begin{bmatrix} 0, 0, 0 \end{bmatrix}^\intercal$.

From \eqref{eq:bbar} and \eqref{eq:pbar}, we form the normalized network input $\mathbf{\bar{X}} := \{\mathbf{\bar{x}}_1, \mathbf{\bar{x}}_2, \cdots, \mathbf{\bar{x}}_n\}$, where each $\mathbf{\bar{x}}_i \in \mathbb{R}^7$ is defined
\begin{align}
	\mathbf{\bar{x}}_i := \begin{bmatrix}
	\mathbf{\bar{b}}_i ^\intercal , \mathbf{\bar{p}}_i ^\intercal
	\end{bmatrix} ^\intercal.
	\label{eq:xibar}
\end{align}

\vspace{2mm}
\noindent \textbf{Normalized Network Loss}. 
A straightforward loss for learning to estimate object depth is direct prediction, i.e.,
\begin{align}
\mathcal{L}_{\text{Abs}}(\textbf{W}) :=  Z_n - f_{\text{Abs}}(\mathbf{X}, \mathbf{W}),
\label{eq:depthloss}
\end{align}
where $\textbf{W}$ are the trainable network parameters, $Z_n$ is the ground truth object depth at $\mathbf{p}_n$, and $f_{\text{Abs}} \in \mathbb{R}$ is the predicted absolute object depth. 
For the input $\mathbf{X}$ in \eqref{eq:depthloss}, we use the $\mathbf{\bar{b}}_i$ bounding box format \eqref{eq:bbar} and make each camera position relative to final prediction position $\mathbf{{p}}_n = \begin{bmatrix} 0, 0, 0 \end{bmatrix}^\intercal$.

To use dimensionless input $\mathbf{\bar{X}}$ \eqref{eq:xibar}, we modify \eqref{eq:depthloss} as
\begin{align}
\mathcal{L}_{\text{Rel}}(\textbf{W}) := \frac{Z_n}{\left\| \mathbf{p}_n - \mathbf{p}_1 \right\|} - f_{\text{Rel}}(\mathbf{\bar{X}}, \mathbf{W}),
\label{eq:relloss}
\end{align} 
where $\frac{Z_n}{\left\| \mathbf{p}_n - \mathbf{p}_1 \right\|}$ is the ground truth object depth at $\mathbf{p}_n$ made dimensionless by its relation to the overall camera movement range and $f_{\text{Rel}} \in \mathbb{R}$ is the corresponding relative depth prediction.
To use this dimensionless relative output at inference, we simply multiply $f_{\text{Rel}}$ by $\left\| \mathbf{p}_n - \mathbf{p}_1 \right\|$ to find $Z_n$.

As we will show in Section~\ref{sec:results}, by using a normalized and dimensionless input-loss formulation \eqref{eq:relloss}, DBox can predict object depth across domains with vastly different image resolutions and camera movement ranges.

\begin{figure} [t!]
	\centering
	\includegraphics[width=0.475\textwidth]{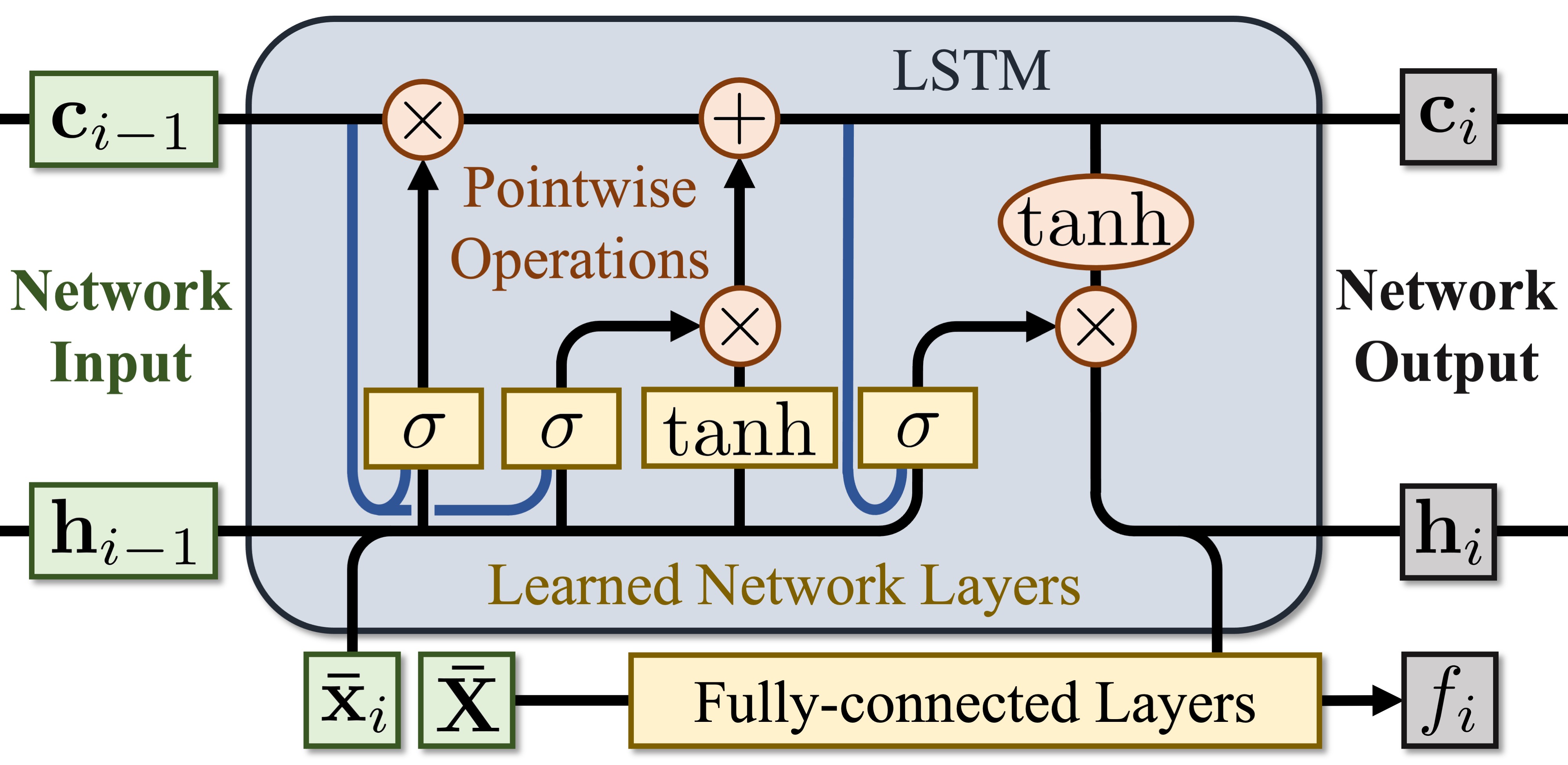}
	\caption{ \textbf{DBox Network}. 
		DBox is a Long Short-Term Memory network with sequential input ({\color{green}green}), five pointwise operations ({\color{red}red}), five learned components ({\color{yellow}yellow}), three $\sigma$-gate peephole connections ({\color{blue}blue}), and a sequential depth output ({\color{gray}gray}).
	}
	\label{fig:dbox}
\end{figure}

\vspace{2mm}
\noindent \textbf{Network Architecture}. 
The DBox architecture is shown in Figure~\ref{fig:dbox}.
DBox is a modified Long Short-Term Memory (LSTM) RNN \cite{HoSc97} with peephole connections for each $\sigma$-gate \cite{GeSc00} and fully-connected (FC) layers that use each LSTM output ($\mathbf{h}_i$) within the context of all $n$ observations ($\mathbf{\bar{X}}$) to predict object depth ($f_i$).
Using initial conditions $\mathbf{c}_{0}, \mathbf{h}_{0} = \mathbf{0}$, each intermediate input ($\mathbf{c}_{i-1}, \mathbf{h}_{i-1},\mathbf{\bar{x}}_i$) and output ($\mathbf{c}_{i}, \mathbf{h}_{i},f_i$) is unique.
DBox operates sequentially across all $n$ observations to predict final object depth $f_n$.

We design the network to maximize depth prediction performance with low GPU requirements.
The LSTM hidden state capacity is 128 (i.e., $\mathbf{c}_{i}, \mathbf{h}_{i} \in \mathbb{R}^{128}$).
Outside the LSTM, $\mathbf{h}_{i}$ is fed into the first of six FC layers, each with 256 neurons, ReLU activation, and $\mathbf{\bar{X}} \in \mathbb{R}^{7n}$ concatenated to their input.
The output of the sixth FC layer is fed to a fully-connected neuron, $f_i \in \mathbb{R}$, the depth output.
Using a modern workstation and GPU (GTX 1080 Ti), a full DBox forward pass of $1 \times 10^{3}$ 10-observation examples uses 677 \textrm{MiB} of GPU memory and takes $4.3 \times 10^{-3}$ seconds.

\subsection{Depth from Motion and Detection Dataset}
\label{sec:odmd}

To train and evaluate our DBox design from Section~\ref{sec:dbox}, we introduce the ODMD dataset.
ODMD enables DBox to learn detection-based depth estimation from any combination of $x,y,z$ camera motion.
Furthermore, although the current ODMD benchmark focuses on robotics, our general ODMD framework can be configured for new applications.

\vspace{2mm}
\noindent \textbf{Generating Motion and Detection Training Data}. 
To generate ODMD training data, we start with the $n$ random camera movements ($\mathbf{p}_i$ \eqref{eq:pi}) in each training example $\mathbf{X}$. 
Given a minimum and maximum movement range as configurable parameters, $\Delta \mathbf{p}_{\text{min}}$ and $\Delta \mathbf{p}_{\text{max}}$, we define a uniform random variable for the overall camera movement as
\begin{align}
 \Delta \mathbf{p} := \mathbf{p}_n - \mathbf{p}_1
	\sim \mathcal{U} [\Delta \mathbf{p}_{\text{min}}, \Delta \mathbf{p}_{\text{max}}] \circ \mathbf{k},
	\label{eq:deltap}
\end{align}
where $\Delta \mathbf{p} \in \mathbb{R}^3$ is the camera movement from $\mathbf{p}_1$ to $\mathbf{p}_n$ and
$\mathbf{k} \in \{-1,1 \}^3$ uses a Rademacher distribution to randomly assign the movement direction of each axis.
As in \eqref{eq:depthloss}, we choose $\mathbf{{p}}_n = \begin{bmatrix} 0, 0, 0 \end{bmatrix}^\intercal \implies \mathbf{p}_1 = -\Delta \mathbf{p}$ in \eqref{eq:deltap}.
For the $1 < i < n$ intermediate camera positions, we use $\mathbf{p}_i \sim \mathcal{U} [\mathbf{p}_1, \mathbf{p}_n]$
then sort the collective $\mathbf{p}_i$ values along each axis so that movement from $\mathbf{p}_1$ to $\mathbf{p}_n$ is monotonic.

After finding the camera movement, we generate a random object with a random initial 3D position. 
To make each object unique, we randomize its physical width and height each as $W, H \sim \mathcal{U} [s_{\text{min}}, s_{\text{max}}]$, where $s_{\text{min}}, s_{\text{max}}$ are configurable.
Using parameters $Z_{1\text{min}}, Z_{1\text{max}}$, we similarly randomize the object's initial depth as $Z_1 \sim \mathcal{U} [Z_{1\text{min}}, Z_{1\text{max}}]$.
To ensure the object is within view, we use $Z_1$ to adjust the bounds of the object's random center position $X_1,Y_1$. Thus,
\begin{align}
X_1 \sim &~\mathcal{U} [X_{1\text{min}} (Z_1), X_{1\text{max}} (Z_1)] \nonumber \\
Y_1 \sim &~\mathcal{U} [Y_{1\text{min}} (Z_1), Y_{1\text{max}} (Z_1)],
\label{eq:X1}
\end{align}
and $\begin{bmatrix}X_1, Y_1, Z_1 \end{bmatrix}^\intercal$ are the object's 3D camera-frame coordinates at camera position $\mathbf{p}_1$.
For completeness, we provide a detailed derivation of \eqref{eq:X1} in the supplementary material.

After finding size and position, we project the object's bounding box onto the camera's image plane at all $n$ camera positions.
As in \eqref{eq:zj}, assuming a static object, all motion is from the camera.
Thus, the object's position at each $\mathbf{p}_i$ is
\begin{align}
	\begin{bmatrix}X_i, Y_i, Z_i \end{bmatrix}^\intercal  = \begin{bmatrix}X_1, Y_1, Z_1 \end{bmatrix}^\intercal - (\mathbf{p}_i - \mathbf{p}_1).
	\label{eq:objecti}
\end{align}
In other words, object motion in the camera frame is equal and opposite to the camera motion itself.
Using $X_i, Y_i, Z_i, W, H$ in \eqref{eq:x} and \eqref{eq:w}, we find bounding box image coordinates $x_i, y_i, w_i, h_i$ \eqref{eq:xi} at all $n$ camera positions, which completes the random training example $\mathbf{X}$.

As a final adjustment for greater variability, because the range of final object position $\begin{bmatrix}X_n, Y_n, Z_n \end{bmatrix}^\intercal$ is greater than that of the initial position $\begin{bmatrix}X_1, Y_1, Z_1 \end{bmatrix}^\intercal$, we randomly reverse the order of all observations in $\mathbf{X}$ with probability 0.5 (i.e., $\{\mathbf{x}_1, \mathbf{x}_2, \cdots, \mathbf{x}_n\}$ changes to $\{\mathbf{x}_n, \mathbf{x}_{n-1}, \cdots, \mathbf{x}_1\}$).



\vspace{2mm}
\noindent \textbf{Learning which Observations to Trust}. 
We add perturbations to ODMD-generated data as an added challenge.
As we will show in Section~\ref{sec:results}, this decision improves the performance of ODMD-trained networks in real applications with object detection and camera movement errors.

For perturbations that cause camera movement errors, we modify \eqref{eq:pi} by adding noise to each $\mathbf{p}_{i}$ for  $1<i\leq n$ as
\begin{align}
\mathbf{p}_{pi} := \mathbf{p}_{i} + \begin{bmatrix} p_{Xi}, p_{Yi}, p_{Zi} \end{bmatrix}^\intercal, ~ p_{Xi}, p_{Yi}, p_{Zi}\sim \mathcal{N}(0,\sigma^2),
\label{eq:pip}
\end{align}
where $\mathbf{p}_{pi}$ is the perturbed version of the camera position $\mathbf{p}_{i}$ and $\mathcal{N}(0,\sigma^2)$ is a Gaussian distribution with $\mu=0$, $\sigma=1 \times 10^{-2}$ that is uniquely sampled for each $p_{Xi}, p_{Yi}, p_{Zi}$.
Note that $\sigma$ can be configured to reflect the anticipated magnitude of errors for a specific application domain.

We use two types of perturbations for object detection errors.
First, similar to \eqref{eq:pip}, we modify \eqref{eq:bbar} by adding noise ($\sigma=1 \times 10^{-3}$) to each bounding box $\mathbf{\bar{b}}_i$ for $1 \leq i\leq n$ as
\begin{align}
\mathbf{\bar{b}}_{pi} := \mathbf{\bar{b}}_{i} + \begin{bmatrix} p_{xi}, p_{yi}, p_{wi}, p_{hi} \end{bmatrix}^\intercal \nonumber \\ 
p_{xi}, p_{yi}, p_{wi}, p_{hi}\sim \mathcal{N}(0,\sigma^2).
\label{eq:xip}
\end{align}
For the second perturbation, we randomly replace one $\mathbf{\bar{b}}_i$ with a \textit{completely} different bounding box with probability 0.1.
This random replacement synthesizes intermittent detections of the wrong object, and the probability of replacement can be configured for a specific application.
Ground truth labels remain the same when we use perturbations.


\vspace{2mm}
\noindent \textbf{ODMD Validation and Test Sets}. 
We introduce four ODMD validation and test sets using robot experiments and simulated data with various levels of perturbations.
This establishes a repeatable benchmark for ablative studies and future methods.
All examples include $n=10$ observations.

The robot experiment data evaluates object depth estimation on a physical platform using object detection on real-world objects.
We collect data with camera movement using a Toyota Human Support Robot (HSR).
HSR has an end effector-mounted wide-angle grasp camera, a 4-DOF arm on a torso with prismatic and revolute joints, and a differential drive base \cite{UiYamaguchi2015,HSR_journal}.
Using HSR's full kinematics, we collect sets of 480$\times$640 grasp-camera images across randomly sampled camera movements ($\Delta \mathbf{p}$ \eqref{eq:deltap}) with target objects in view (see example in Figure~\ref{fig:overview}).

\begin{figure} [t]
	\centering
	\includegraphics[width=0.475\textwidth]{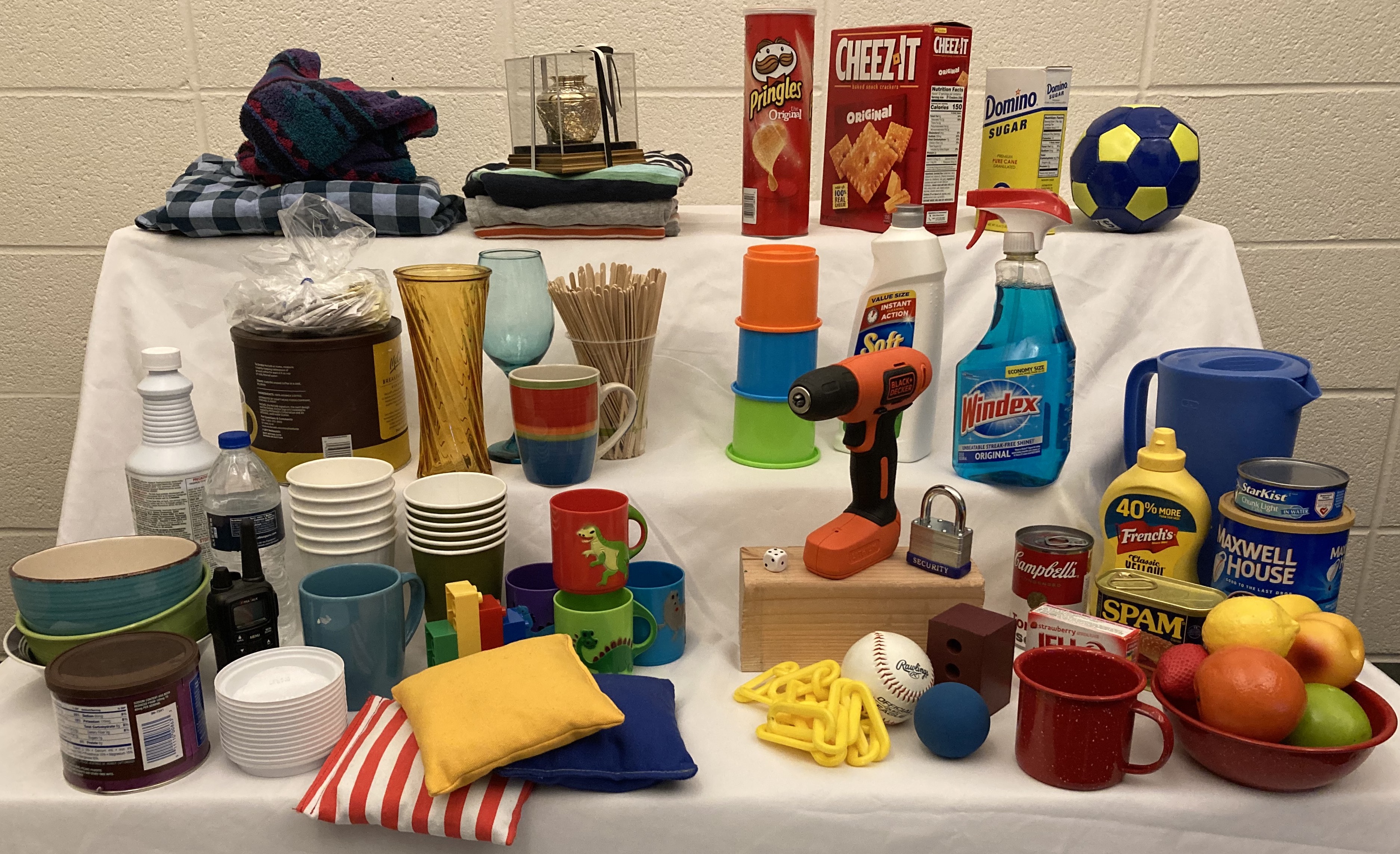}
	\caption{ \textbf{Robot Experiment Objects}. We use 30 custom objects in the Validation Set (left) and 30 YCB objects in the Test Set (right).
		Object size spans from 16~\textrm{mm} (die) to 280~\textrm{mm} (shirt).
	}
	\label{fig:robot_objects}
\end{figure}

For the robot experiment objects, we use 30 custom household object for the Validation Set and 30 YCB objects \cite{YCB} for the Test Set (see Figure~\ref{fig:robot_objects}) across six different scenes.
We detect objects using Faster R-CNN \cite{ReEtAl15}, which we fine-tune on each set of objects using custom annotation images outside of the validation and test sets.
The camera movement range ($\Delta \mathbf{p}$) varies between $\begin{bmatrix}-0.34, -0.33, -0.43 \end{bmatrix}^\intercal$ to $\begin{bmatrix}0.33, 0.36, 0.37 \end{bmatrix}^\intercal$ \textrm{m}, and the final object depth ($Z_n$), which we measure manually, varies between 0.11-1.64 \textrm{m}.
Altogether, we generate 5,400 robot object depth estimation examples (2,400 validation and 3,000 test), and we show a few challenging examples in the supplementary material.

We generate a set of normal and two types of perturbation-based data for simulated objects.
The Normal Set approximates HSR's operation without any object detection or camera movement errors.
Using ODMD's random data-generation framework, we choose configurable parameters: $s_{\text{min}}=0.01, s_{\text{max}}=0.175$ \textrm{m} for object size; $\Delta \mathbf{p}_{\text{min}} = \begin{bmatrix} 0,0,0.05 \end{bmatrix}^\intercal, \Delta \mathbf{p}_{\text{max}} = \begin{bmatrix} 0.25,0.175,0.325 \end{bmatrix}^\intercal$ \textrm{m} for camera movement \eqref{eq:deltap}; and $Z_{1\text{min}}=0.55, Z_{1\text{max}}=1$ \textrm{m} for initial object depth.
For the camera model, we use factory-provided intrinsics for HSR's 480$\times$640 grasp camera, specifically, $f_x, f_y= 205.5, c_x=320.5, c_y=240.5$.
This configuration satisfies camera view constraints in \eqref{eq:X1}.

The two perturbation-based sets approximate HSR's operation with only object detection or camera movement errors.
For the Camera Motion Perturbation Set, we use the same configuration as the Normal Set with camera movement noise added \eqref{eq:pip}.
For the Object Detection Perturbation Set, we use the normal configuration with bounding box noise and random box replacement \eqref{eq:xip}.
We generate 5,400 depth estimation examples for each of the simulated object configurations (2,400 validation and 3,000 test).

\vspace{2mm}
\noindent \textbf{ODMD Training Configurations for DBox}. 
To test the efficacy of new concepts and establish best practices, we train DBox using multiple ODMD training configurations.
The first configuration is the same as ODMD's \textbf{N}ormal validation and test \textbf{S}et (DBox$_{\text{NS}}$) but trains on continuously-generated data.
Similarly, the second configuration (DBox$_p$) is based on both perturbation sets and trains on new data with camera movement errors \eqref{eq:pip}, object detection errors \eqref{eq:xip}, and random replacement.

DBox$_{\text{NS}}$ and DBox$_p$ both learn to predict object depth relative to camera movement using the dimensionless loss $\mathcal{L}_{\text{Rel}}$ \eqref{eq:relloss}.
Alternatively, the third configuration (DBox$_{\text{Abs}}$) learns to predict absolute object depth using $\mathcal{L}_{\text{Abs}}$ \eqref{eq:depthloss}.
DBox$_{\text{Abs}}$ uses the same training perturbations as DBox$_p$.

We train each network using a batch size of 512 randomly-generated training examples with $n=10$ observations per prediction. 
We train each network for $1 \times 10^{7}$ iterations using the Adam Optimizer \cite{adam14} with a $1\times10^{-3}$ learning rate, which takes approximately 2.8 days using a single GPU (GTX 1080 Ti).
We also train alternate versions of DBox$_p$ for $1 \times 10^{6}$ (DBox$_p^{1\textrm{M}}$) and $1 \times 10^{5}$ iterations (DBox$_p^{100\textrm{K}}$), which takes 6.7 and 0.7 hours respectively.


For compatibility with the ODMS benchmark \cite{GrCo20}, we train three final configurations with \textit{only} $z$-axis camera motion and more general camera parameters.
The first configuration (DBox$_{p}^z$) is the same as DBox$_p$ but with $\Delta \mathbf{p}_{\text{max}} = \begin{bmatrix} 0,0,0.4625 \end{bmatrix}^\intercal$ \textrm{m} and $f_x, f_y= 240.5$.
The other configurations are the same as DBox$_{p}^z$ but either remove perturbations (DBox$_{\text{NS}}^z$) or use loss $\mathcal{L}_{\text{Abs}}$ (DBox$_{\text{Abs}}^z$).
Because single-axis camera movement is simple, we train each configuration for only $1 \times 10^{4}$ iterations, which takes less than four minutes.

\section{Experimental Results}
\label{sec:results}

\subsection{Setup}

For the first set of experiments, we evaluate DBox on the new ODMD benchmark in Section~\ref{sec:odmdres}.
We find the number of training iterations for the results using the best overall validation performance, which we check at every hundredth of the total training iterations.
Like the ODMS benchmark \cite{GrCo20}, we evaluate each method using the mean percent error for each test set, which we calculate for each example as
\begin{align}
\text{Percent Error} = \left| \frac{Z_n - \hat{Z}_n}{Z_n} \right| \times 100 \%,
\label{eq:percent}
\end{align}
where $Z_n$ and $\hat{Z}_n$ are ground truth and predicted object depth at final camera position $\mathbf{p}_n$.
Finally, for all of our results, if an object is not detected at position $\mathbf{p}_i$, we use the bounding box from the nearest position with a detection.

We compare DBox to state-of-the-art methods on the object segmentation-based ODMS benchmark in Section~\ref{sec:odmsres}.
Because ODMS provides only $z$-axis camera motion and segmentation, we preprocess inputs for DBox.
First, we set all $x$- and $y$-axis camera inputs to zero.
Second, we create bounding boxes around the object segmentation masks using the minimum and maximum $x,y$ pixel locations.
For instances where a mask consists of multiple disconnected fragments (e.g., from segmentation errors), we use the segment minimizing $\frac{\text{center offset}}{\text{\# of pixels}}$, which generally keeps the bounding box to the intended target rather than including peripheral errors.
Finally, as with ODN \cite{GrCo20}, we find the number of training iterations for each test set using the best corresponding validation performance, which we check at every hundredth of the total training iterations.


We test a new application using DBox with a camera phone in Section~\ref{sec:phone}.
We take sets of ten pictures with target objects in view, and we estimate the camera movement between images using markings on the ground (e.g., sections of sidewalk).
Like the ODMD Robot Set in Section~\ref{sec:odmd}, the ground truth object depth is manually measured, and we detect objects using Faster R-CNN fine-tuned on separate training images.
Overall, we generate 46 camera phone object depth estimation examples across a variety of settings.

\subsection{ODMD Dataset}
\label{sec:odmdres}

\setlength{\tabcolsep}{2.25pt} 
\begin{table} [t]
	\centering
	\caption{\textbf{Object Depth via Motion and Detection Results}.
	ODMD inputs have full $x,y,z$ motion and bounding boxes.}
	\footnotesize
	\begin{tabular}{| l | c | c | c | c | c | c | c | c |}
		\hline	& & &	\multicolumn{5}{c|}{ Mean Percent Error  \eqref{eq:percent} }   \\	\cline{4-8}
		\multicolumn{1}{|c|}{} & \multicolumn{1}{c|}{Object}  &\multicolumn{1}{c|}{}  & \multicolumn{1}{c|}{}  & \multicolumn{2}{c|}{Perturb} & \multicolumn{1}{c|}{} &   \\ \cline{5-6}
		\multicolumn{1}{|c|}{Config.}  &	\multicolumn{1}{c|}{Depth} & \multicolumn{1}{c|}{Train}  &\multicolumn{1}{c|}{}  & \multicolumn{1}{c|}{Camera}	&	\multicolumn{1}{c|}{Object} & \multicolumn{1}{c|}{} & All  \\
		\multicolumn{1}{|c|}{ID} &	\multicolumn{1}{c|}{Method}  & \multicolumn{1}{c|}{Data} &	\multicolumn{1}{c|}{Norm.}  & \multicolumn{1}{c|}{Motion}	&	\multicolumn{1}{c|}{Detect.} & \multicolumn{1}{c|}{Robot} & Sets \\	\hline 	\hline
DBox$_{p}$	&	$f_{\text{Rel}}$ \eqref{eq:relloss}	&	Perturb	&	1.7	&	2.5	&	2.5	& \bf	11.2	& \bf	4.5	\\
\rowcolor{rowgray}	DBox$_{\text{Abs}}$	&	$f_{\text{Abs}}$ \eqref{eq:depthloss}	&	Perturb	&	1.1	& \bf	2.1	& \bf	1.8	&	13.3	&	4.6	\\
DBox$_p^{1\textrm{M}}$	&	$f_{\text{Rel}}$ \eqref{eq:relloss}	&	Perturb	&	1.7	&	2.6	&	2.6	&	11.5	&	4.6	\\
\rowcolor{rowgray}	DBox$_p^{100\textrm{K}}$	&	$f_{\text{Rel}}$ \eqref{eq:relloss}	&	Perturb	&	2.2	&	3.0	&	3.0	&	11.7	&	5.0	\\
DBox$_{\text{NS}}$	&	$f_{\text{Rel}}$ \eqref{eq:relloss}	&	Normal	&	0.5	&	3.9	&	6.4	&	12.5	&	5.8	\\
\rowcolor{rowgray}	Box$_\text{LS}$	&	$\hat{Z}_n$  \eqref{eq:axb}	&	N/A	& \bf	0.0	&	4.5	&	21.6	&	21.2	&	11.8	\\
DBox$_{p}^z$	&	$f_{\text{Rel}}$ \eqref{eq:relloss}	&	$z$-axis &	12.9	&	12.5	&	15.0	&	22.0	&	15.6	\\
		\hline
	\end{tabular}
	\label{tab:odmd}
\end{table}

We provide ablative ODMD Test Set results in Table~\ref{tab:odmd}.
We evaluate the five full-motion DBox configurations, $z$-axis only DBox$_{p}^z$, and analytical solution Box$_{\text{LS}}$ \eqref{eq:axb}.
All results use $n=10$ observations (same in Sections~\ref{sec:odmsres}-\ref{sec:phone}), and ``All Sets" is an aggregate score across all test sets.

DBox$_p$ has the best result for the Robot Set and overall.
DBox$_{\text{Abs}}$ comes in second overall and has the best result for the Perturb Camera Motion and Object Detection sets.
However, DBox$_{\text{Abs}}$ is the least accurate of any full-motion DBox configuration on the Robot Set.
Essentially, DBox$_{\text{Abs}}$ gets a performance boost from a camera movement range- and depth-based prior (i.e., $\mathbf{p}_i,\mathbf{X}$ and $f_{\text{Abs}}$ in \eqref{eq:depthloss}) at the cost of generalization to other domains with different movement and depth profiles.
In Section~\ref{sec:odmsres}, this trend becomes most apparent for DBox$_{\text{Abs}}^z$ in the driving domain.

Analytical solution Box$_\text{LS}$ is perfect on the error-free Normal Set but performs worse relative to other methods on sets with input errors, especially object detection errors.
Similarly, DBox$_{\text{NS}}$ has a great result on the Normal Set but is the least accurate overall of the full-motion DBox configurations. 
Intuitively, DBox$_{\text{NS}}$ trains without perturbations, so it is more susceptible to input errors.
In Section~\ref{sec:odmsres}, this trend becomes most apparent for DBox$_{\text{NS}}^z$ in the robot domain, which has many object segmentation errors \cite{GrCo20}.

DBox$_{p}^z$ uses only $z$-axis camera motion and is the least accurate overall.
Thus, for applications with full $x,y,z$ motion, this result shows the importance of training on examples with motion parallax and using all three camera motion inputs to determine depth.
We provide more results comparing depth estimation cues in the supplementary material.

We find that we can train the state-of-the-art DBox$_p$ much faster if needed.
DBox$_p^{1\textrm{M}}$ trains for one tenth the iterations of DBox$_p$ and performs almost as well.
Going further, training for one hundredth the iterations is more of a trade off, as DBox$_p^{100\textrm{K}}$ has an overall 11\% increase in error.

\subsection{ODMS Dataset}
\label{sec:odmsres}

We provide comparative results for the ODMS benchmark \cite{GrCo20} in Table~\ref{tab:odms}.
We evaluate the three $z$-axis DBox configurations, Box$_{\text{LS}}$, and current state-of-the-art methods.

DBox$_{p}^z$ achieves the best Robot, Driving, and overall result on the segmentation-based ODMS benchmark despite training only on detection-based data.
Furthermore, while the second best method, ODN$_{\ell r}$, takes 2.6 days to train on segmentation masks \cite{GrCo20}, DBox$_{p}^z$ trains in under four minutes by using a simpler bounding box-based input \eqref{eq:xibar}.

\setlength{\tabcolsep}{2.75pt} 
\begin{table} [t]
	\centering
	\caption{\textbf{Object Depth via Motion and Segmentation Results}.
	ODMS inputs have $z$-axis motion and binary segmentation masks.}
	\footnotesize
	\begin{tabular}{| l | c | c| c | c | c | c |}
		\hline	\multicolumn{1}{|c|}{Object} & \multicolumn{1}{c|}{} &	\multicolumn{5}{c|}{ Mean Percent Error  \eqref{eq:percent} }  \\	\cline{3-7}
		\multicolumn{1}{|c|}{Depth}	& \multicolumn{1}{c|}{Vision} & \multicolumn{2}{c|}{Simulated} &  \multicolumn{1}{c|}{} & \multicolumn{1}{c|}{} & All \\ \cline{3-4}
		\multicolumn{1}{|c|}{Method} & \multicolumn{1}{c|}{Input} 	& \multicolumn{1}{c|}{Normal} & \multicolumn{1}{c|}{Perturb} & \multicolumn{1}{c|}{Robot}  & \multicolumn{1}{c|}{Driving} & Sets \\	\hline 			
		\multicolumn{7}{c}{Learning-based Methods} \\
		\hline
DBox$_{p}^z$	&	Detection	&	11.8	&	20.3	& \bf	11.5	& \bf	24.8	& \bf	17.1	\\	
\rowcolor{rowgray}	ODN$_{\ell r}$  \cite{GrCo20}	&	Segmentation	&	8.6	&	17.9	&	13.1	&	31.7	&	17.8	\\	
ODN$_{\ell p}$  \cite{GrCo20}	&	Segmentation	&	11.1	& \bf	13.0	&	22.2	&	29.0	&	18.8	\\	
\rowcolor{rowgray}	ODN$_\ell$ \cite{GrCo20}	&	Segmentation	& \bf	8.3	&	18.2	&	19.3	&	30.1	&	19.0	\\	
DBox$_{\text{NS}}^z$	&	Detection	&	9.2	&	31.6	&	39.3	&	37.3	&	29.3	\\	
\rowcolor{rowgray}	DBox$_{\text{Abs}}^z$	&	Detection	&	21.3	&	25.5	&	20.4	&	53.1	&	30.1	\\	\hline \multicolumn{7}{c}{Analytical Methods} \\ \hline
Box$_\text{LS}$~\eqref{eq:axb}	&	Detection	&	13.7	&	36.6	& \bf	17.6	& \bf	33.3	& \bf	25.3	\\	
\rowcolor{rowgray}	VOS-DE \cite{GrFlCo20}	&	Segmentation	& \bf	7.9	& \bf	33.6	&	32.6	&	36.0	&	27.5	\\	
\hline
	\end{tabular}
	\label{tab:odms}
\end{table}
\begin{figure}[t!]
	\centering
	\input{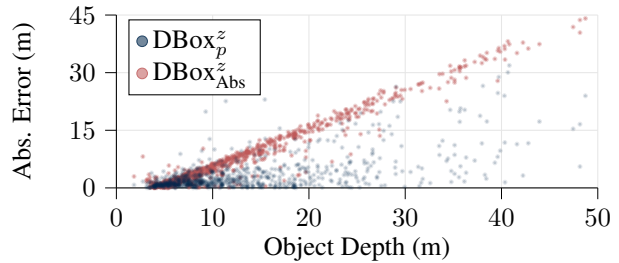}
	\caption{\textbf{Absolute Error vs. Object Depth for the Driving Set}.}
	\label{fig:error}
\end{figure}
\begin{figure*} [t!]
	\centering
	\includegraphics[width=0.8\textwidth]{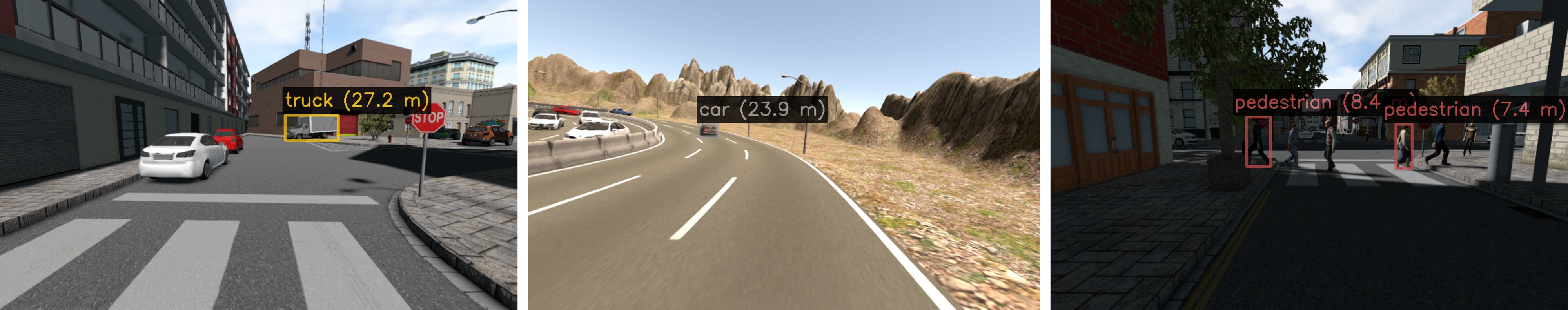}
	\caption{ \textbf{Example Depth Results for Driving}.
	}
	\label{fig:synthia}
\end{figure*}
\begin{figure*} [t!]
	\centering
	\includegraphics[width=0.8\textwidth]{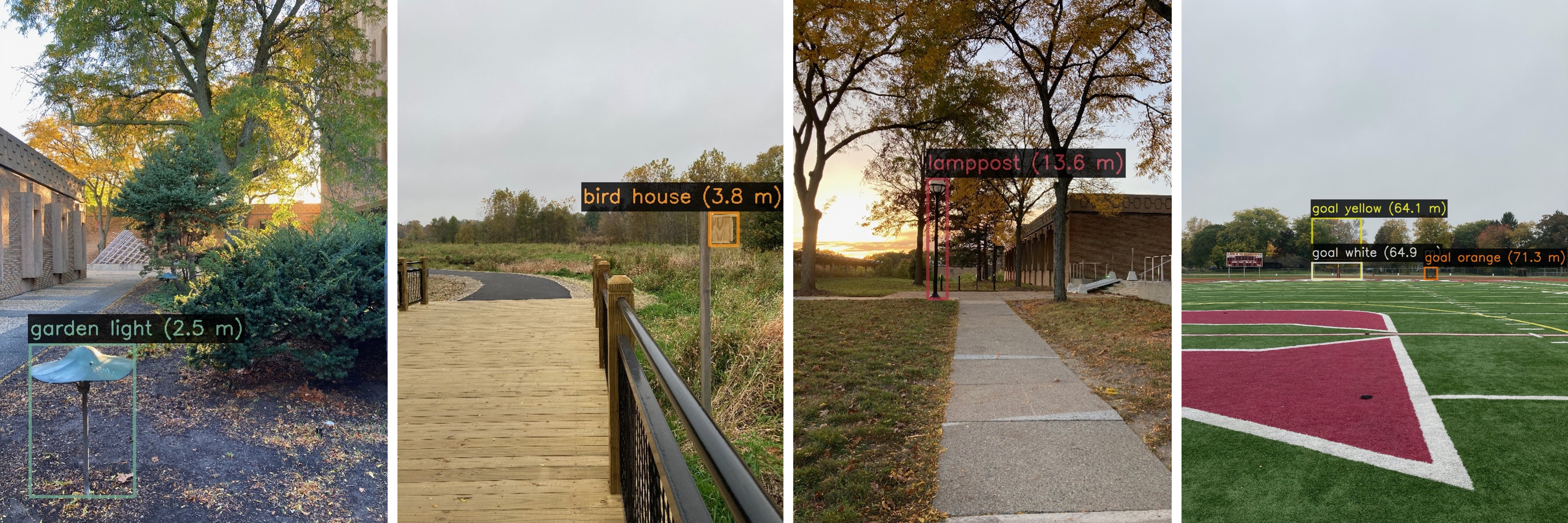}
	\caption{ \textbf{Example Depth Results using a Camera Phone}. 
	}
	\label{fig:phone}
\end{figure*}

For analytical methods, Box$_{\text{LS}}$ achieves the best Robot, Driving, and overall result.
Box$_{\text{LS}}$ improves robustness by using width- and height-based changes in scale \eqref{eq:axb}, generating twice as many solvable equations as VOS-DE \cite{GrFlCo20}, but performs worse on Simulated sets due to mask conversion.

State-of-the-art DBox$_{p}^z$ is significantly more reliable across applications than other DBox configurations.
DBox$_{\text{NS}}^z$ is the least accurate Robot Set method, and DBox$_{\text{Abs}}^z$ is the least accurate Driving Set method, while DBox$_{p}^z$ is the best method for both.
Considering the DBox configuration differences, we attribute DBox$_{p}^z$'s success to using three types of perturbations during training \eqref{eq:pip}-\eqref{eq:xip} and a dimensionless loss ($\mathcal{L}_{\text{Rel}}$ \eqref{eq:relloss}).
Notably, perturbations from previous work are less effective (ODN$_{\ell p}$, ODN$_{\ell}$  \cite{GrCo20}).

To compare DBox$_{p}^z$ and DBox$_{\text{Abs}}^z$ on the Driving Set, we plot the absolute error ($| Z_n - \hat{Z}_n |$) vs.~ground truth object depth ($Z_n$) in Figure~\ref{fig:error}.
Because DBox$_{\text{Abs}}^z$ predictions are biased toward short distances within its learned prior ($f_{\text{Abs}}$ \eqref{eq:depthloss}), DBox$_{\text{Abs}}^z$ errors offset directly with object depth.
Alternatively, DBox$_{p}^z$ predictions are dimensionless ($f_{\text{Rel}}$ \eqref{eq:relloss}) and less affected by long distances.
We show four Driving results for DBox$_{p}^z$ in Figure~\ref{fig:synthia}, which have an error of 0.95~\textrm{m} (truck), 1.5~\textrm{m} (car), and 1.0~\textrm{m} and -0.59~\textrm{m} (pedestrians).

\subsection{Depth using a Camera Phone}
\label{sec:phone}



We check the performance of DBox$_{p}^z$ on all camera phone examples at every hundredth of the total training iterations, which achieves a mean percent error \eqref{eq:percent} of 6.7 in under four minutes of training.
Overall, DBox$_{p}^z$'s robustness to imprecise camera movement and generalization to a camera with vastly different parameters than the training model is highly encouraging.
We show six results in Figure~\ref{fig:phone}, which have an error of 0.30~\textrm{m} (garden light), 0.31~\textrm{m} (bird house), 0.16~\textrm{m} (lamppost), and 0.10~\textrm{m}, 0.85~\textrm{m}, and 7.3~\textrm{m} (goal yellow, goal white, and goal orange).


\subsection{Final Considerations for Implementation}

For applications with full $x,y,z$ camera motion, DBox$_{p}$ is state-of-the-art.
Alternatively, for applications with only $z$-axis motion, DBox$_{p}^z$ is state-of-the-art.
Thus, although each network is broadly applicable, we suggest configuring ODMD training to match an application's camera movement profile for best performance.
Finally, because DBox's prediction speed is negligible for many applications, we suggest combining predictions over many permutations of collected data to improve accuracy \cite[Section~6.2]{GrCo20}.

\section{Conclusions}

We develop an analytical model and multiple approaches to estimate object depth using uncalibrated camera motion and bounding boxes from object detection.
To train and evaluate methods in this new problem space, we introduce the \textbf{O}bject \textbf{D}epth via \textbf{M}otion and \textbf{D}etection (ODMD) benchmark dataset, which includes mobile robot experiments using a single RGB camera to locate objects.
ODMD training data are extensible, configurable, and include three types of perturbations typical of camera motion and object detection.
Furthermore, we show that training with perturbations improves performance in real-world applications.

Using the ODMD dataset, we train the first network to estimate object depth from motion and detection.
Additionally, we develop a generalized representation of bounding boxes, camera movement, and relative depth prediction, which we show to improve general applicability across vastly different domains.
Using a single ODMD-trained network with object detection \textit{or} segmentation, we achieve state-of-the-art results on existing driving and robotics benchmarks and accurately estimate the depth of objects using a camera phone.
Given the network's reliability across domains and real-time operation, we find our approach to be a viable tool to estimate object depth in mobile applications. 


\vspace{2mm}
\noindent \textbf{Acknowledgements}. 
Toyota Research Institute provided funds to support this work.

{\small
\bibliographystyle{ieee_fullname}
\bibliography{GrCoCVPR21}
}

\clearpage

	\begin{center}
		\Large\textbf{Supplementary Material:}\\
		\Large\textbf{Depth from Camera Motion and Object Detection}
	\end{center}


\section*{Depth from Motion Parallax \& Detection} 

Building off of the model from Section~\ref{sec:model}, if there is $x$- or $y$-axis camera motion between observations, we can solve for object dept using corresponding changes in bounding box location (e.g., Box$_1$ to Box$_3$, Figure~\ref{fig:3D_box}).

To start, we account for any incidental depth-based changes in scale by reformulating \eqref{eq:fxw} as 
\begin{align}
Z_j =  Z_i \Big( \frac{w_i}{w_j} \Big).
\label{eq:zjmp}
\end{align}
Next, we use \eqref{eq:zjmp} in \eqref{eq:x} to relate bounding box center coordinate $x_j$ to corresponding 3D object coordinate $X_{j}$ as
\begin{align}
x_j = \frac{f_x X_{j}}{Z_j} + c_x  = \frac{f_x X_{j}}{ Z_i \Big( \frac{w_i}{w_j} \Big)} + c_x \nonumber \\
\implies (x_j - c_x) \Big( \frac{w_i}{w_j} \Big) = \frac{f_x X_{j}}{ Z_i}.
\label{eq:xj}
\end{align}
Given a static object, changes in lateral object position $X_{j}$ occur only from changes in camera position $\text{C}_{Xi}$ \eqref{eq:pi}. Thus,
\begin{align}
X_{j} - X_{i} = -(\text{C}_{Xj} - \text{C}_{Xi}).
\label{eq:xcj}
\end{align}

Finally, using \eqref{eq:xj} and \eqref{eq:xcj}, we can solve for $Z_i$ by comparing two observations $i,j$ with motion parallax as
\begin{align}
(x_j - c_x) \Big( \frac{w_i}{w_j} \Big) - (x_i - c_x) = \frac{f_x (X_{j} - X_{i})}{ Z_i} \nonumber \\
\implies Z_i = \frac{f_x (\text{C}_{Xi} - \text{C}_{Xj})}{ (x_j - c_x) \Big( \frac{w_i}{w_j} \Big) - (x_i - c_x) }.
\label{eq:zimp}
\end{align}
Notably, \eqref{eq:zimp} can also be derived using vertical motion as
\begin{align}
Z_i = \frac{f_y (\text{C}_{Yi} - \text{C}_{Yj})}{ (y_j - c_y) \Big( \frac{w_i}{w_j} \Big) - (y_i - c_y) },
\label{eq:zivmp}
\end{align}
and scale measure $\frac{h_i}{h_j}$ can replace $\frac{w_i}{w_j}$ in \eqref{eq:zimp} or \eqref{eq:zivmp}.
Also, if there is no $z$-axis camera motion (i.e., $Z_i=Z_j$), then $\frac{w_i}{w_j} = \frac{h_i}{h_j} = 1$ and we can simplify \eqref{eq:zimp} and \eqref{eq:zivmp} as
\begin{align}
	Z_i = \frac{f_x (\text{C}_{Xi} - \text{C}_{Xj})}{ x_j - x_i } = \frac{f_y (\text{C}_{Yi} - \text{C}_{Yj})}{ y_j - y_i }.
\end{align}

\section*{Comparison of Depth Estimation Cues} 

We provide ODMD Test Set results in Table~\ref{tab:odmdmp} to compare solutions using different depth estimation cues. 
We evaluate three different analytical solutions that use single cues and DBox$_{\text{NS}}$, which uses full $x,y,z$ motion.

For the motion parallax solution, we use the average of the lateral \eqref{eq:zimp} and vertical \eqref{eq:zivmp} motion parallax solutions, using scale measure $\frac{w_i}{w_j}$ in \eqref{eq:zimp} and $\frac{h_i}{h_j}$ for \eqref{eq:zivmp}.
Notably, this is a two-observation solution, so we use the end point observations of each example, i.e., $i=n=10$ and $j=1$.
For comparison, we similarly evaluate a two-observation, optical expansion-based solution, which uses the average of \eqref{eq:zi} when using $\frac{w_i}{w_j}$ and $\frac{h_i}{h_j}$ for the end point observations.

Motion parallax performs the best overall for the two-observation solutions in Table~\ref{tab:odmdmp}.
The optical expansion solution performs surprising well with camera motion perturbations but much worse on the test sets with object detection errors (i.e., Perturb Object Detection and Robot).
Both solutions are perfect on the error-free Normal Set.

The Box$_\text{LS}$ solution, which uses optical expansion over all $n$ observations, significantly outperforms both two-observation solutions, especially on test sets with object detection errors.
Thus, for applications with real-world detection (e.g., the Robot Set), we find that incorporating many observations is more beneficial than choosing between optical expansion or motion parallax with fewer observations.

DBox$_{\text{NS}}$, using all $n$ observations and full $x,y,z$ motion, performs the best overall and on all test sets with any kind of input errors.
Admittedly, some error mitigation likely results from DBox$_{\text{NS}}$ using a probabilistic learning-based method.
Still, DBox$_{\text{NS}}$ trains on ideal data without any input errors, so DBox$_{\text{NS}}$ predictions are based on an ideal model, just like the analytical methods.
Accordingly, we postulate that DBox$_{\text{NS}}$'s improvement over Box$_\text{LS}$ is primarily the result of learning full $x,y,z$ motion features, which are more reliable than a single depth cue (e.g., DBox$_{p}^z$ in Table~\ref{tab:odmd}).

Although our analytical model in Section~\ref{sec:model} and current methods focus on $x,y,z$ camera motion, 
adding rotation as an additional depth estimation cue is an area of future work.
Nonetheless, our state-of-the-art results on the ODMS Driving Set in Table~\ref{tab:odms} do include examples with camera rotation from vehicle turning  \cite[Section~5.2]{GrCo20} (Figure~\ref{fig:synthia}, center).
Finally, as a practical consideration for robotics applications, motion planners using our current approach can simply incorporate rotation after estimating depth.




\setlength{\tabcolsep}{2pt} 
\begin{table} [t]
	\centering
	\caption{\textbf{ODMD Results for Various Depth Estimation Cues}.}
	\footnotesize
	\begin{tabular}{| l | c | c | c | c | c | c | c |}
		\hline	& &	\multicolumn{5}{c|}{ Mean Percent Error  \eqref{eq:percent} }   \\	\cline{3-7}
		\multicolumn{1}{|c|}{} & \multicolumn{1}{c|}{Object}   & \multicolumn{1}{c|}{}  & \multicolumn{2}{c|}{Perturb} & \multicolumn{1}{c|}{} &   \\ \cline{4-5}
		\multicolumn{1}{|c|}{Analytical Depth}  &	\multicolumn{1}{c|}{Depth}  &\multicolumn{1}{c|}{}  & \multicolumn{1}{c|}{Camera}	&	\multicolumn{1}{c|}{Object} & \multicolumn{1}{c|}{} & All  \\
		\multicolumn{1}{|c|}{Estimation Cue} &	\multicolumn{1}{c|}{Method}  & 	\multicolumn{1}{c|}{Norm.}  & \multicolumn{1}{c|}{Motion}	&	\multicolumn{1}{c|}{Detect.} & \multicolumn{1}{c|}{Robot} & Sets \\	\hline
				\multicolumn{7}{c}{Learning-based Methods} \\ \hline
		Full $x,y,z$ Motion	&	DBox$_{\text{NS}}$ \eqref{eq:relloss}	&	0.5	&	\bf	3.9	&	\bf	6.4	&	\bf	12.5	&	\bf	5.8	\\ \hline
		\multicolumn{7}{c}{Analytical Methods} \\ \hline
		 Optical Expansion	&	Box$_\text{LS}$  \eqref{eq:axb}	&	\bf	0.0	&	4.5	&	21.6	&	21.2	&	11.8	\\
		\rowcolor{rowgray}Motion Parallax &	$Z_n$ \eqref{eq:zimp}-\eqref{eq:zivmp}	&	\bf 0.0	&	33.9	&	51.6	&	65.6	&	37.8	\\
		Optical Expansion	&	$Z_n$ \eqref{eq:zi} &	\bf	0.0	&	5.2	&	80.9	&	124.1	&	52.5	\\
		\hline
	\end{tabular}
	\label{tab:odmdmp}
\end{table}

\section*{Camera-based Constraints on Generated Data}

When generating new ODMD training data in Section~\ref{sec:odmd}, we consider the full camera model to ensure that generated 3D objects and their bounding boxes are within the camera's field of view.
To derive this constraint, we first note that the center of a bounding box \eqref{eq:xi} is within view if $x_i \in~[0,W_I], y_i~\in~[0,H_I]$, where $W_I, H_I$ are the image width and height. 
Using \eqref{eq:x}, we represent these constraints for 3D camera-frame coordinates $X_i,Y_i$ as
\begin{align}
0 \leq x_i = \frac{f_x X_i}{Z_i} + &~ c_x \leq W_I, ~0 \leq y_i = \frac{f_y Y_i}{Z_i} + c_y \leq H_I\nonumber \\ 
\implies &~ \frac{-c_x Z_i}{f_x} \leq X_i \leq \frac{(W_I - c_x) Z_i}{f_x} \label{eq:xlim}\\
\implies &~ \frac{-c_y Z_i}{f_y} \leq  Y_i \leq \frac{(H_I - c_y) Z_i}{f_y}.  \label{eq:ylim}
\end{align}

We also consider constraints based on the maximum object size $s_{\text{max}}$ and camera movement range $\Delta \mathbf{p}_{\text{max}}$ \eqref{eq:deltap}. 
We use $\Delta \mathbf{p}_{\text{max}}$ by defining it in terms of its components parts as
\begin{align}
	\Delta \mathbf{p}_{\text{max}} := \begin{bmatrix} \Delta \text{C}_{X\text{max}}, \Delta \text{C}_{Y\text{max}}, \Delta \text{C}_{Z\text{max}} \end{bmatrix}^\intercal.
	\label{eq:pmax}
\end{align} 
Then, using $s_{\text{max}}$, $\Delta \mathbf{p}_{\text{max}}$, and the initial object position $\begin{bmatrix}X_1, Y_1, Z_1 \end{bmatrix}^\intercal$ \eqref{eq:X1}, we update the constraints in \eqref{eq:xlim} as
\begin{align}
\frac{-c_x (Z_1 - \Delta \text{C}_{Z\text{max}})}{f_x} \leq &~ X_1 - \Delta \text{C}_{X\text{max}} - \frac{s_\text{max}}{2} \leq  \nonumber \\ 
X_1 + \Delta \text{C}_{X\text{max}} + \frac{s_\text{max}}{2} \leq &~ \frac{(W_I - c_x) (Z_1 - \Delta \text{C}_{Z\text{max}})}{f_x}, \label{eq:xlimgen}
\end{align}
and, equivalently for height, update constraints in \eqref{eq:ylim} as
\begin{align}
\frac{-c_y (Z_1 - \Delta \text{C}_{Z\text{max}})}{f_y} \leq &~ Y_1 - \Delta \text{C}_{Y\text{max}} - \frac{s_\text{max}}{2} \leq  \nonumber \\ 
Y_1 + \Delta \text{C}_{Y\text{max}} + \frac{s_\text{max}}{2} \leq &~ \frac{(H_I - c_y) (Z_1 - \Delta \text{C}_{Z\text{max}})}{f_y}, \label{eq:ylimgen}
\end{align}
where $Z_1 - \Delta \text{C}_{Z\text{max}}$ accounts for camera approach to the object, $\Delta \text{C}_{X\text{max}}$ and $\Delta \text{C}_{Y\text{max}}$ account for lateral and vertical camera movement, and $\frac{s_\text{max}}{2}$ accounts for object width and height.
Because \eqref{eq:xlimgen}-\eqref{eq:ylimgen} use the maximum camera movement range and object size, they guarantee, first, \eqref{eq:xlim}-\eqref{eq:ylim} are satisfied for all $n$ object positions $\begin{bmatrix}X_i, Y_i, Z_i \end{bmatrix}^\intercal$ and, second, all corresponding bounding boxes are in view.

Given $X_1,Y_1$, we can find the lower bound for the initial object depth ($Z_{1\text{min}}$) by replacing $Z_1$ with $Z_{1\text{min}}$ in \eqref{eq:xlimgen} and \eqref{eq:ylimgen} to find
\begin{align}
Z_{1\text{min}} \geq \Delta \text{C}_{Z\text{max}} + \text{max}\Bigg( \Big( \frac{f_x}{c_x} \Big) \Big( \frac{s_\text{max}}{2} + \Delta \text{C}_{X\text{max}} - X_1 \Big), \nonumber \\
\Big( \frac{f_y}{c_y} \Big) \Big( \frac{s_\text{max}}{2} + \Delta \text{C}_{Y\text{max}} - Y_1 \Big), \nonumber \\
\Big( \frac{f_x}{W_I -c_x} \Big) \Big( \frac{s_\text{max}}{2} + \Delta \text{C}_{X\text{max}} + X_1 \Big), \nonumber \\
\Big( \frac{f_y}{H_I -c_y} \Big) \Big( \frac{s_\text{max}}{2} + \Delta \text{C}_{Y\text{max}} + Y_1 \Big) \Bigg).
\label{eq:z1lb}
\end{align}
In other words, given an object's center position and maximum size, the minimum viewable depth is constrained by the closest image boundary after camera movement.
Note that there is no equivalent upper bound for $Z_{1\text{max}}$.

Given $Z_1$, similar to \eqref{eq:z1lb}, we can find the lower and upper bounds for $X_1,Y_1$ in \eqref{eq:X1} using \eqref{eq:xlimgen} and \eqref{eq:ylimgen} to find
\begin{align}
X_{1\text{min}} &~ \geq \Big( \frac{c_x}{f_x} \Big)( \Delta \text{C}_{Z\text{max}} -Z_1 ) + \Delta \text{C}_{X\text{max}} + \frac{s_\text{max}}{2}  \nonumber \\
Y_{1\text{min}} &~ \geq \Big( \frac{c_y}{f_y} \Big)( \Delta \text{C}_{Z\text{max}} -Z_1 ) + \Delta \text{C}_{Y\text{max}} + \frac{s_\text{max}}{2}  \nonumber \\
X_{1\text{max}} &~ \leq \Big( \frac{W_I -c_x}{f_x} \Big)( Z_1 - \Delta \text{C}_{Z\text{max}}) - \Delta \text{C}_{X\text{max}} - \frac{s_\text{max}}{2} \nonumber \\
Y_{1\text{max}} &~ \leq \Big( \frac{H_I -c_y}{f_y} \Big)( Z_1 - \Delta \text{C}_{Z\text{max}}) - \Delta \text{C}_{Y\text{max}} - \frac{s_\text{max}}{2}. \label{eq:xmax}
\end{align}

When generating ODMD training data in Section~\ref{sec:odmd}, we cannot select the $Z_{1\text{min}}$ constraint simultaneously with the $X_{1\text{min}},Y_{1\text{min}},X_{1\text{max}},Y_{1\text{max}}$ constraints in \eqref{eq:xmax}.
 Alternatively, we choose a $Z_{1\text{min}}$ value greater than the lower bound for $X_1,Y_1=0$ in \eqref{eq:z1lb}, then randomly select $Z_1 \sim \mathcal{U} [Z_{1\text{min}}, Z_{1\text{max}}]$ for each training example. 
Once $Z_1$ is randomly determined, we use \eqref{eq:xmax} to find 
 \begin{align}
X_{1\text{min}}(Z_1) &~ = \Big( \frac{c_x}{f_x} \Big) ( \Delta \text{C}_{Z\text{max}} -Z_1 ) + \Delta \text{C}_{X\text{max}} + \frac{s_\text{max}}{2} \nonumber \\
Y_{1\text{min}}(Z_1) &~ = \Big( \frac{c_y}{f_y} \Big)( \Delta \text{C}_{Z\text{max}} -Z_1 ) + \Delta \text{C}_{Y\text{max}} + \frac{s_\text{max}}{2}  \nonumber \\
X_{1\text{max}}(Z_1) &~ = \Big( \frac{W_I -c_x}{f_x} \Big)( Z_1 - \Delta \text{C}_{Z\text{max}}) - \Delta \text{C}_{X\text{max}} \nonumber \\ &  
~~~~~~~~~~~~~~~~~~~~~~~~~~~~~~~~~~~~~~~~~~~~~~~~~~~~-\frac{s_\text{max}}{2} \nonumber \\
Y_{1\text{max}}(Z_1) &~ = \Big( \frac{H_I -c_y}{f_y} \Big)( Z_1 - \Delta \text{C}_{Z\text{max}}) - \Delta \text{C}_{Y\text{max}}  \nonumber \\&  ~~~~~~~~~~~~~~~~~~~~~~~~~~~~~~~~~~~~~~~~~~~~~~~~~~~~-\frac{s_\text{max}}{2},
\label{eq:X1min}
 \end{align}
 which is the exact solution we use in \eqref{eq:X1}.
 Notably, in absence of making adjustments for the specific object size or camera movement range of each example, \eqref{eq:X1min} provides the greatest range of initial positions that also guarantees the object is in view for all $n$ observations.
 Finally, \eqref{eq:X1min} is linear, so we vectorize it for large batches of training examples.

\setlength{\tabcolsep}{4.75pt}
\begin{table}[t!]
	\centering
	\caption{\textbf{Detailed ODMD Results}.} 
	\footnotesize
	\begin{tabular}{|l|c|c|c|c|c|}
		\hline
		\multicolumn{1}{|c|}{Object} & \multicolumn{5}{c|}{Percent Error \eqref{eq:percent}} \\
		\cline{2-6}
		\multicolumn{1}{|c|}{Depth} &  & & \multicolumn{2}{c|}{Range}  & \multicolumn{1}{c|}{Standard}  \\
		\cline{4-5}
		\multicolumn{1}{|c|}{Method} & Mean & Median & Minimum & Maximum & \multicolumn{1}{c|}{Deviation}   \\
		\hline
		\multicolumn{6}{c}{Normal Set} \\
		\hline
DBox$_{p}$	& 1.73 & 0.96 & 0.0002 & 48.21 &  2.90 \\
\rowcolor{rowgray}	DBox$_{\text{Abs}}$	& 1.11 & 0.82 & 0.0004 & 21.10 & 1.19 \\
DBox$_{\text{NS}}$	& 0.54 & 0.38 & 0.0001 & 8.68 & 0.63 \\
\rowcolor{rowgray}	Box$_\text{LS}$	& \bf 0.00 & \bf 0.00 & \bf 0.0000 & \bf 0.00 & \bf 0.00 \\
DBox$_{p}^z$ & 12.89 & 8.54 & 0.0062 & 80.74 & 13.23 \\
		\hline
		\multicolumn{6}{c}{Perturb Camera Motion Set} \\
		\hline
DBox$_{p}$	& 2.45 & 1.86 & 0.0008 & 23.61 & 2.28 \\
\rowcolor{rowgray}	DBox$_{\text{Abs}}$	& \bf  2.05 & \bf  1.55 & \bf  0.0002 & \bf  19.45 & \bf 1.96 \\
DBox$_{\text{NS}}$	& 3.91 & 2.93 & 0.0021 & 47.94 &  3.82 \\
\rowcolor{rowgray}	Box$_\text{LS}$	& 4.47 & 3.13 & 0.0007 & 43.02 & 4.57 \\
DBox$_{p}^z$ & 12.48 & 8.42 & 0.0025 & 74.18 &  12.18 \\ \hline
		\multicolumn{6}{c}{Perturb Object Detection Set} \\
\hline
DBox$_{p}$ & 2.54 & 1.54 & 0.0020 & 45.94 & 3.39 \\
\rowcolor{rowgray}	DBox$_{\text{Abs}}$	 & \bf 1.75 & \bf 1.26 & 0.0007 & \bf 19.51 & \bf  1.81 \\
DBox$_{\text{NS}}$ & 6.35 & 1.98 & 0.0005 & 415.68 & 19.56 \\
\rowcolor{rowgray}	Box$_\text{LS}$	& 21.60 & 8.90 & \bf 0.0003 & 158.04 & 28.27 \\
DBox$_{p}^z$ & 15.00 & 9.83 & 0.0189 & 296.31 & 16.93 \\ \hline
		\multicolumn{6}{c}{Robot Set} \\
\hline
DBox$_{p}$	& \bf 11.17 & 8.31 & 0.0022 & 253.02 & \bf 13.94 \\
\rowcolor{rowgray}	DBox$_{\text{Abs}}$	& 13.29 & 9.44 & 0.0024 & \bf 223.76 & 14.90 \\
DBox$_{\text{NS}}$	& 12.47 & \bf 8.11 & 0.0092 & 656.85 & 25.03 \\
\rowcolor{rowgray}	Box$_\text{LS}$	& 21.23 & 12.17 & \bf 0.0010 & 262.48 & 26.92 \\
DBox$_{p}^z$ & 21.96 & 14.64 & 0.0099 & 342.40 & 26.39 \\
		\hline
	\end{tabular}
	\label{tab:detailodmd}
\end{table}

\setlength{\tabcolsep}{4.75pt}
\begin{table}[t!]
	\centering
	\caption{\textbf{Detailed ODMS Results}.}
	\footnotesize
	\begin{tabular}{|l|c|c|c|c|c|}
		\hline
		\multicolumn{1}{|c|}{Object} & \multicolumn{5}{c|}{Percent Error \eqref{eq:percent}} \\
		\cline{2-6}
		\multicolumn{1}{|c|}{Depth} &  & & \multicolumn{2}{c|}{Range}  & \multicolumn{1}{c|}{Standard}  \\
		\cline{4-5}
		\multicolumn{1}{|c|}{Method} & Mean & Median & Minimum & Maximum & \multicolumn{1}{c|}{Deviation}   \\
		\hline
		\multicolumn{6}{c}{Normal Set} \\
		\hline
		DBox$_{p}^z$ & 11.82 & 8.17 & 0.0049 & 167.80 & 12.42 \\
		\rowcolor{rowgray}	Box$_\text{LS}$ & 13.66 & 10.49 & \bf 0.0025 & \bf 137.39 & 11.97 \\
		DBox$_{\text{NS}}^z$ & \bf 9.20 & \bf 6.69 & 0.0048 & 146.79 & \bf 9.55 \\
		\rowcolor{rowgray}	DBox$_{\text{Abs}}^z$  & 21.31 & 11.98 & 0.0119 & 451.54 & 33.95 \\
		\hline
		\multicolumn{6}{c}{Perturb Set} \\
		\hline
		DBox$_{p}^z$ & \bf 20.34 & 15.25 & \bf 0.0008 & 220.46 & \bf 19.73 \\ 
\rowcolor{rowgray}	Box$_\text{LS}$ & 36.62 & 27.76 & 0.0050 & \bf 141.85 & 30.06 \\
DBox$_{\text{NS}}^z$ & 31.55 & 19.95 & 0.0205 & 644.55 & 48.03 \\
\rowcolor{rowgray}	DBox$_{\text{Abs}}^z$ & 25.49 & \bf 15.12 & 0.0033 & 265.11 & 30.68 \\ \hline
		\multicolumn{6}{c}{Robot Set} \\
		\hline
		DBox$_{p}^z$ & \bf 11.45 & 6.29 & 0.0061 & 418.41 & \bf 23.81 \\
\rowcolor{rowgray}	Box$_\text{LS}$ & 17.62 & 9.15 & \bf 0.0011 & 390.12 & 34.22 \\
DBox$_{\text{NS}}^z$ & 39.25 & \bf 5.97 & 0.0082 & 8778.45 & 310.94 \\
\rowcolor{rowgray}	DBox$_{\text{Abs}}^z$ & 20.36 & 10.28 & 0.0033 & \bf 358.86 & 32.80 \\
		\multicolumn{6}{c}{Driving Set} \\
		\hline
		DBox$_{p}^z$ & \bf 24.84 & \bf 18.99 & 0.0323 & \bf 213.93 & \bf 22.83 \\
\rowcolor{rowgray}	Box$_\text{LS}$ & 33.29 & 26.50 & 0.1783 & 294.91 & 31.10 \\
DBox$_{\text{NS}}^z$ & 37.31 & 21.43 & \bf 0.0108 & 613.14 & 55.75 \\
\rowcolor{rowgray}	DBox$_{\text{Abs}}^z$ & 53.13 & 55.89 & 0.0878 & 296.88 & 26.65 \\
		\hline
	\end{tabular}
	\label{tab:detailodms}
\end{table}

\setlength{\tabcolsep}{4.75pt}
\begin{table}[t!]
	\centering
	\caption{\textbf{Detailed ODMD Results (Absolute Error)}.} 
	\footnotesize
	\begin{tabular}{|l|c|c|c|c|c|}
		\hline
		\multicolumn{1}{|c|}{Object} & \multicolumn{5}{c|}{Absolute Error \eqref{eq:abs}} \\
		\cline{2-6}
		\multicolumn{1}{|c|}{Depth} &  & & \multicolumn{2}{c|}{Range}  & \multicolumn{1}{c|}{Standard}  \\
		\cline{4-5}
		\multicolumn{1}{|c|}{Method} & Mean & Median & Minimum & Maximum & \multicolumn{1}{c|}{Deviation}   \\
		\hline
		\multicolumn{6}{c}{Normal Set (\textrm{cm})} \\
		\hline
		DBox$_{p}$ & 1.42 & 0.69 & 0.0002 & 46.57 & 2.62 \\
		\rowcolor{rowgray}	DBox$_{\text{Abs}}$	& 0.87 & 0.59 & 0.0003 & 11.56 & 0.95 \\
		DBox$_{\text{NS}}$	& 0.41 & 0.28 & 0.0001 & 8.97 & 0.53 \\
		\rowcolor{rowgray}	Box$_\text{LS}$	 & \bf 0.00 & \bf 0.00 & \bf 0.0000 & \bf 0.00 & \bf 0.00 \\
		DBox$_{p}^z$ & 10.30 & 6.22 & 0.0040 & 77.13 & 11.68 \\
		\hline
		\multicolumn{6}{c}{Perturb Camera Motion Set (\textrm{cm})} \\
		\hline
		DBox$_{p}$ & 1.93 & 1.38 & 0.0005 & 20.84 & 1.93 \\
		\rowcolor{rowgray}	DBox$_{\text{Abs}}$	& \bf 1.63 & \bf 1.13 & \bf 0.0001 & \bf 17.17 & \bf 1.65 \\
		DBox$_{\text{NS}}$ & 3.04 & 2.16 & 0.0024 & 41.03 & 3.12 \\
		\rowcolor{rowgray}	Box$_\text{LS}$	& 3.44 & 2.37 & 0.0005 & 33.65 & 3.66 \\
		DBox$_{p}^z$ & 9.92 & 6.26 & 0.0022 & 67.80 & 10.57 \\ \hline
		\multicolumn{6}{c}{Perturb Object Detection Set (\textrm{cm})} \\
		\hline
		DBox$_{p}$ & 2.06 & 1.11 & 0.0017 & 42.25 & 3.07 \\
		\rowcolor{rowgray}	DBox$_{\text{Abs}}$ & \bf 1.39 & \bf 0.93 & 0.0005 & \bf 15.91 & \bf 1.55 \\
		DBox$_{\text{NS}}$ & 5.01 & 1.47 & 0.0004 & 281.55 & 15.07 \\
		\rowcolor{rowgray}	Box$_\text{LS}$	& 17.58 & 7.08 & \bf 0.0002 & 121.45 & 23.67 \\
		DBox$_{p}^z$ & 11.81 & 7.12 & 0.0089 & 146.10 & 13.43 \\ \hline
		\multicolumn{6}{c}{Robot Set (\textrm{cm})} \\
		\hline
		DBox$_{p}$ & \bf 8.08 & 5.79 & 0.0012 & 260.28 & 12.06 \\
		\rowcolor{rowgray}	DBox$_{\text{Abs}}$	& 8.83 & 6.71 & 0.0018 & \bf 55.83 & \bf 7.87 \\
		DBox$_{\text{NS}}$ & 9.23 & \bf 5.57 & 0.0045 & 579.77 & 23.51 \\
		\rowcolor{rowgray}	Box$_\text{LS}$	& 14.49 & 8.63 & \bf 0.0007 & 197.56 & 17.70 \\
		DBox$_{p}^z$ & 14.65 & 10.29 & 0.0089 & 161.98 & 14.69 \\
		\hline
	\end{tabular}
	\label{tab:odmdabs}
\end{table}

\setlength{\tabcolsep}{4.75pt}
\begin{table}[t!]
	\centering
	\caption{\textbf{Detailed ODMS Results (Absolute Error)}.}
	\footnotesize
	\begin{tabular}{|l|c|c|c|c|c|}
		\hline
		\multicolumn{1}{|c|}{Object} & \multicolumn{5}{c|}{Absolute Error \eqref{eq:abs}} \\
		\cline{2-6}
		\multicolumn{1}{|c|}{Depth} &  & & \multicolumn{2}{c|}{Range}  & \multicolumn{1}{c|}{Standard}  \\
		\cline{4-5}
		\multicolumn{1}{|c|}{Method} & Mean & Median & Minimum & Maximum & \multicolumn{1}{c|}{Deviation}   \\
		\hline
		\multicolumn{6}{c}{Normal Set (\textrm{cm})} \\
		\hline
		DBox$_{p}^z$ & 3.65 & 2.70 & 0.0020 & \bf 30.77 & \bf 3.66 \\
		\rowcolor{rowgray}	Box$_\text{LS}$ & 4.74 & 3.43 & \bf 0.0006 & 55.77 & 4.81 \\
		DBox$_{\text{NS}}^z$ &\bf  2.98 & \bf 2.14 & 0.0008 & 82.58 & 3.74 \\
		\rowcolor{rowgray}	DBox$_{\text{Abs}}^z$ & 5.57 & 3.98 & 0.0059 & 50.57 & 5.81 \\
		\hline
		\multicolumn{6}{c}{Perturb Set (\textrm{cm})} \\
		\hline
		DBox$_{p}^z$  & 7.19 & \bf 4.46 & \bf 0.0003 & 77.16 & 8.06 \\ 
		\rowcolor{rowgray}	Box$_\text{LS}$ & 15.17 & 8.30 & 0.0016 & 79.01 & 16.45 \\
		DBox$_{\text{NS}}^z$  & 12.21 & 5.77 & 0.0091 & 295.98 & 23.27 \\
		\rowcolor{rowgray}	DBox$_{\text{Abs}}^z$  & \bf 6.68 & 5.20 & 0.0004 & \bf 37.75 & \bf 5.65 \\ \hline
		\multicolumn{6}{c}{Robot Set (\textrm{cm})} \\
		\hline
		DBox$_{p}^z$ & \bf 3.34 & 1.78 & 0.0013 & 88.89 & \bf 5.94 \\
		\rowcolor{rowgray}	Box$_\text{LS}$ & 5.21 & 2.58 & \bf 0.0005 & 79.71 & 10.54 \\
		DBox$_{\text{NS}}^z$ & 12.06 & \bf 1.71 & 0.0019 & 1634.35 & 84.61 \\
		\rowcolor{rowgray}	DBox$_{\text{Abs}}^z$ & 5.64 & 3.04 & 0.0010 & \bf 70.20 & 7.93 \\ \hline
		\multicolumn{6}{c}{Driving Set (\textrm{m})} \\
		\hline
		DBox$_{p}^z$  & \bf 3.63 & \bf 1.86 & 0.0031 & \bf 37.95 & \bf 5.00 \\
		\rowcolor{rowgray}	Box$_\text{LS}$ & 5.08 & 2.35 & 0.0142 & 58.07 & 8.07 \\
		DBox$_{\text{NS}}^z$ & 5.03 & 2.37 & \bf 0.0004 & 105.97 & 8.43 \\
		\rowcolor{rowgray}	DBox$_{\text{Abs}}^z$ & 9.05 & 5.82 & 0.0046 & 57.60 & 9.24 \\
		\hline
	\end{tabular}
	\label{tab:odmsabs}
\end{table}

\begin{figure*} [t!]
	\centering
	\includegraphics[width=0.505\textwidth]{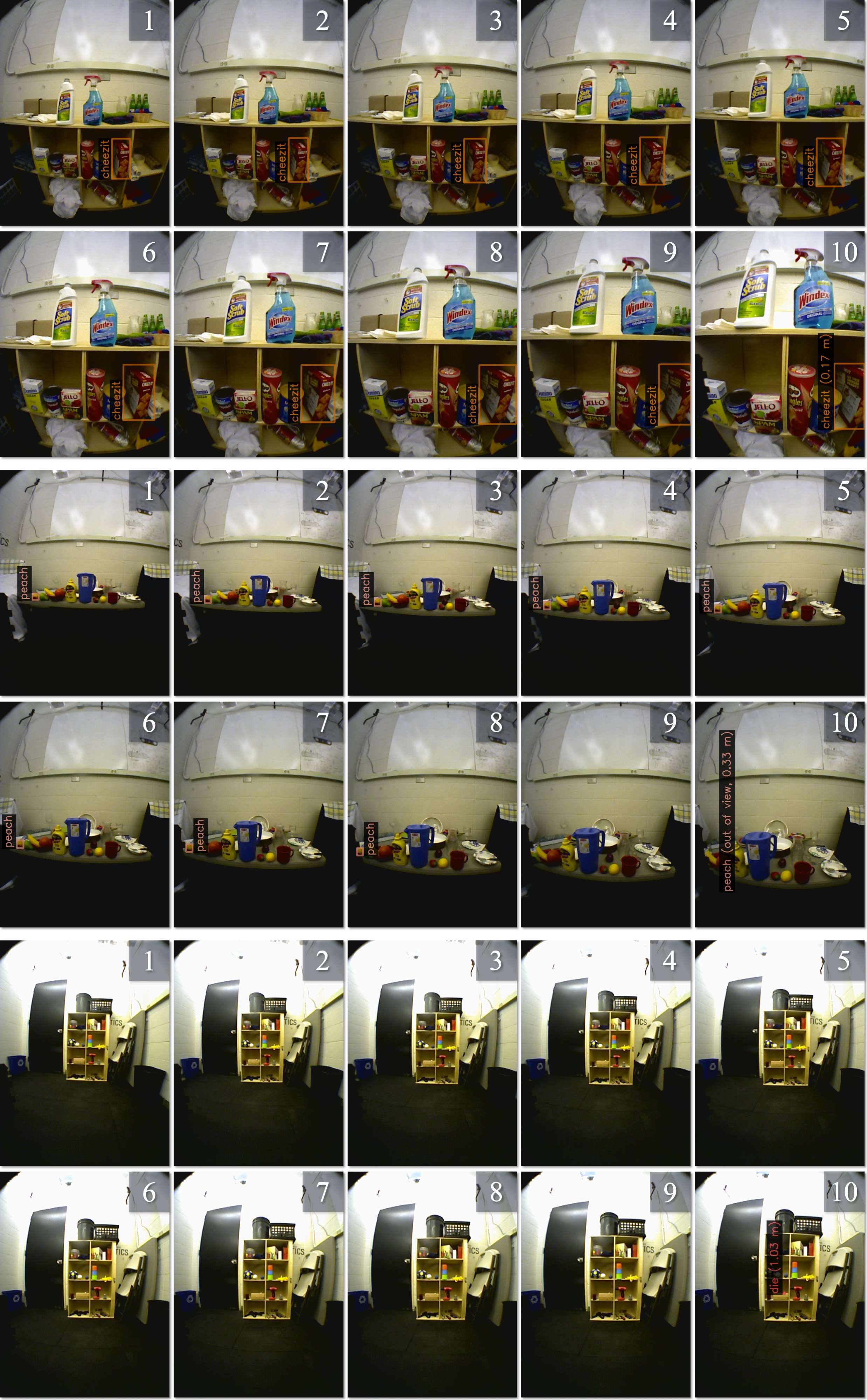}
	\caption{ \textbf{ODMD Robot Test Set Examples} (best viewed in electronic format).
		Every two rows show a ten-observation example progressing from left to right, and we show the ground truth object depth in the final image.
		In the cheezit example (top two rows), the camera perspective changes and the detected object partially leaves view (observations 8-10), causing a distortion to the bounding box shape relative to earlier observations.
		In the peach example (middle two rows), the detected object completely leaves view during the final two observations (9-10), providing no bounding box information at the prediction location.
		Finally, for the 16~\textrm{mm} die example (bottom), the camera starts far away from the small object (1), which is not detected until the camera is closer in the final observation (10).
	}
	\label{fig:odmd}
\end{figure*}

\section*{Detailed ODMD and ODMS Results}

We provide more comprehensive and detailed ODMD and ODMS results in Tables~\ref{tab:detailodmd} and \ref{tab:detailodms}. Specifically, we provide a more precise mean percent error \eqref{eq:percent} and include the median, range, and standard deviation for each test set.

We also provide ODMD and ODMS results for the absolute error in Tables~\ref{tab:odmdabs} and \ref{tab:odmsabs}, which we calculate for each example as
\begin{align}
	\text{Absolute Error} = \left| Z_n - \hat{Z}_n \right|,
	\label{eq:abs}
\end{align}
where $Z_n$ and $\hat{Z}_n$ are ground truth and predicted object depth at final camera position $\mathbf{p}_n$.
Notably, we use percent error \eqref{eq:percent} in the paper to provide a consistent comparison across domains (and examples) with markedly different object depth distances.
For example, the 0.10 \textrm{m} absolute error from Figure~\ref{fig:phone} is a much better result for a camera phone application than it would be for robot grasping.


 \section*{Object Motion Considerations}

The ODMS Driving Set includes moving objects [\cite{GrCo20}, Section~5.2].
On the other hand, our analytical model in Section~\ref{sec:model} assumes static objects.
Nonetheless, DBox$_p^z$ achieves the current state-of-the-art result on the ODMS Driving Set in Table~\ref{tab:odms}. 
We attribute DBox$_p^z$'s success to training with camera movement perturbations \eqref{eq:pip}.
Note that training with these perturbations improves robustness to input errors for the relative distance changes between the camera and object, whether caused by camera motion errors \textit{or} unintended motion of the object itself.
In general, objects that move much less than the camera are not an issue.

 \section*{ODMD Robot Test Set Examples}
 
For the ODMD Robot Test Set, we intentionally select challenging objects and settings that make object detection and depth estimation difficult.
To illustrate this point, we show a few example challenges in Figure~\ref{fig:odmd}.

\end{document}